\documentclass[runningheads]{llncs}

\usepackage{eccv}

\usepackage{multirow}
\definecolor{LightRed}{rgb}{1.0, 0.8, 0.8}
\usepackage{tcolorbox}
\usepackage{listings}
\tcbuselibrary{breakable, skins}

\usepackage{eccvabbrv}

\usepackage{graphicx}
\usepackage{booktabs}

\usepackage[accsupp]{axessibility}  

\usepackage{hyperref}

\usepackage{orcidlink}

\begin{document}

\title{Beyond Absolute Scores: Relative Edit-induced Difference for Generalizable Image Aesthetic Assessment} 

\titlerunning{Relative Edit-Induced Difference for Image Aesthetic Assessment}

\author{
Qifei Jia$^{\star}$\inst{1} \and
Xintong Yao$^{\star}$\inst{1} \and
Yasen Zhang$^{\dagger}$\inst{1} \and
Minghao Li\inst{1} \and
Yajie Chai\inst{1} \and
Qiming Lu\inst{1} \and
Baoyue Shen\inst{1} \and
Runyu Shi\inst{1} \and
Ying Huang\inst{1} \and
Yue Zhang
}

\authorrunning{Q.~Jia, X.~Yao et al.}

\institute{
Xiaomi Corporation, Beijing, China \\
\email{\{jiaqifei1, yaoxintong1, zhangyasen\}@xiaomi.com}
}

\maketitle

\let\thefootnote\relax
\footnotetext{$^{\star}$~Equal contribution. \\$^{\dagger}$~Corresponding author.}

\begin{abstract}
  Traditional Image Aesthetic Assessment (IAA) methods mainly rely on regressing absolute Mean Opinion Scores (MOS). However, such a paradigm overlooks the inherently dynamic nature of human aesthetic perception, which relies on subconscious comparison against implicit visual references. Consequently, the lack of causal reasoning regarding aesthetic differences prevents models from learning generalizable aesthetic principles, thus limiting their generalization across diverse scenarios. In this work, we rethink the IAA task and propose \textbf{R}elative \textbf{E}dit-induced \textbf{D}ifference \textbf{Aes}thetic learning (\textbf{RED-Aes}), a novel framework that leverages controllable image editing models to simulate the human aesthetic reasoning process. Instead of fitting absolute score distributions, RED-Aes explicitly learns the visual factors that drive aesthetic changes. To support this paradigm, we construct the \textbf{RED-20k} dataset, which comprises editing-based image pairs, quantitative aesthetic differences, and Chain-of-Thought (CoT) reasoning. Furthermore, we introduce a three-stage training strategy guided by a relative ranking consistency reward, optimizing the model solely via relative supervision. Extensive experiments demonstrate that RED-Aes achieves state-of-the-art performance on multiple public benchmarks, exhibiting superior generalization capabilities.
\end{abstract}

\section{Introduction}
\label{sec:intro}

Image Aesthetic Assessment (IAA) aims to quantify the aesthetic appeal of images. It underpins a growing range of real-world applications. In computational photography, on-device aesthetic models power best-shot selection and intelligent photo curation on modern smartphones~\cite{talebi2018nima}. Similarly, in the era of AI-generated content (AIGC), aesthetic scores serve as training-data filters~\cite{schuhmann2022laion} and reward signals for aligning text-to-image generative models~\cite{xu2023imagereward}. Although recent works leverage Vision-Language Models (VLMs)~\cite{Bai2025Qwen25VLTR,liu2023visual} for aesthetic assessment~\cite{wu2023q,you2024descriptive,liu2025unlocking,wu2024q}, existing methods typically rely on crowdsourced Mean Opinion Scores (MOS) as supervision to regress absolute scores on limited datasets.

\begin{figure*}[t]
    \centering
    \includegraphics[width=\linewidth]{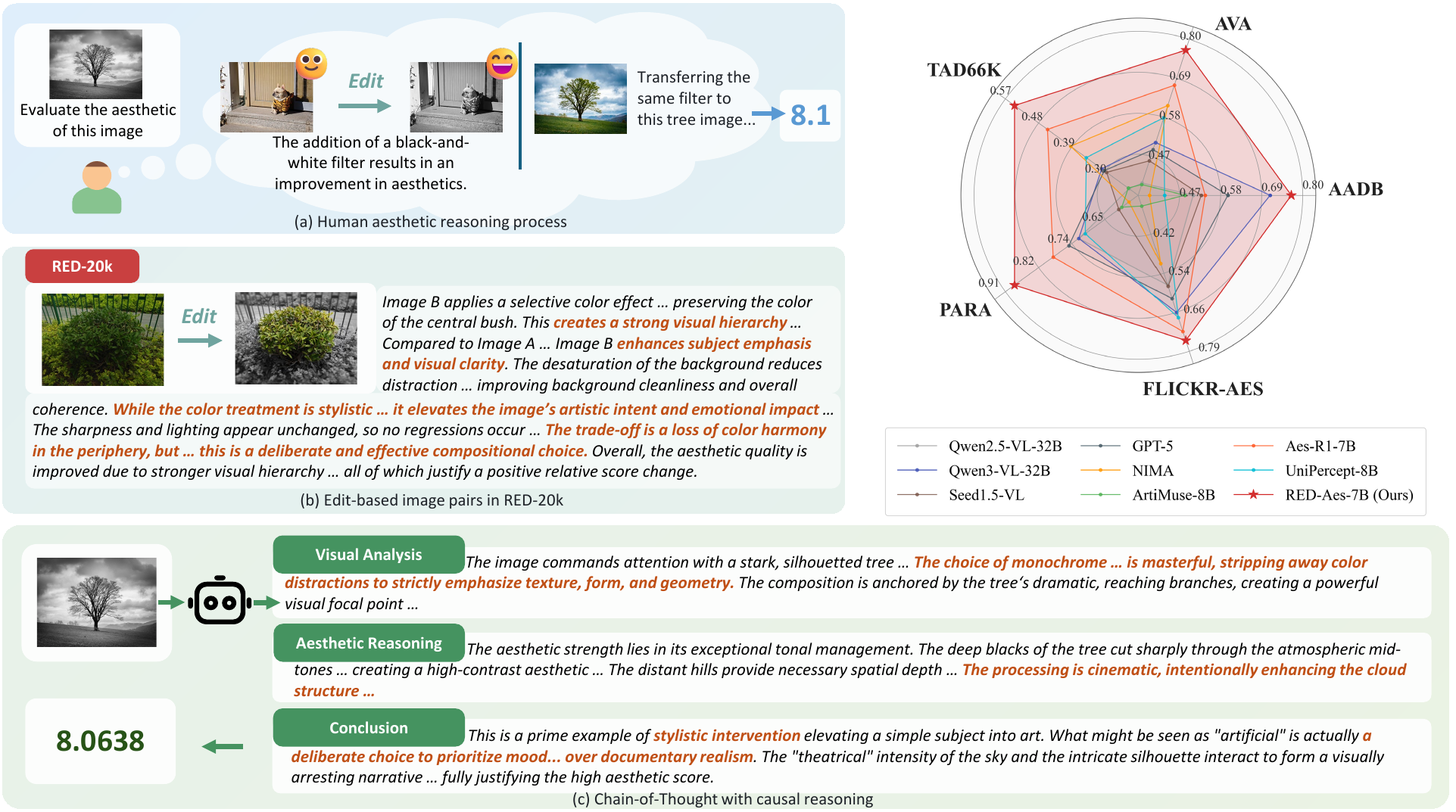}
    \caption{Relative Difference Learning for Aesthetics.
(a) Human aesthetic reasoning process: Illustration of how humans rely on implicit comparison to judge aesthetics.
(b) Edit-based image pairs in RED-20k: An example of source-edit pairs where edit operations induce quantifiable aesthetic shifts.
(c) CoT with causal reasoning: The model produces detailed textual explanations covering visual analysis and aesthetic reasoning to justify the predicted score.}
    \label{fig:intro}
\end{figure*}

This paradigm overlooks a critical aspect of human aesthetic cognition. In practice, a human annotator rarely assigns an absolute score directly but instead relies on a reference anchor. This anchor is either an explicitly provided sample or an implicitly recalled image of similar content~\cite{leder2004model,reber2004processing} which allows the annotator to reason about differences to arrive at a score. While recent VLM-based approaches assess aesthetics from multiple dimensions~\cite{wu2023q,you2024descriptive,liu2025unlocking}, the causal link between reasoning and scores remains weak. Forcing models to fit score distributions without comparative knowledge constrains generalization. Human aesthetic judgment is thus a dynamic comparative process involving causal reasoning about aesthetic differences rather than an instantaneous absolute measurement, as illustrated in \cref{fig:intro}(a).

This insight has been validated in the Image Quality Assessment (IQA) domain where RankIQA~\cite{Liu_2017_ICCV} constructs quality ranking pairs by applying synthetic distortions of varying severity to source images to improve generalization. However, aesthetics is a high-level attribute dependent on content and semantics that cannot be manipulated through simple pixel-level operations. With the advancement of controllable image editing models~\cite{brooks2023instructpix2pix,zhang2023magicbrush,zhao2024ultraedit}, this idea can now be transferred to IAA. Targeted edits such as composition adjustments, color harmony enhancements, or lighting modifications create source-edit pairs (\cref{fig:intro}(b)). The aesthetic differences within these pairs encode explicit causal relationships where specific modifications directly induce quantifiable aesthetic shifts. This mechanism provides far stronger supervision than statistical correlations derived solely from absolute scores.

Building on this insight, we propose Relative Edit-induced Difference Aesthetic learning (RED-Aes), a framework that simulates how humans establish aesthetic concepts through comparison. RED-Aes explicitly learns relative aesthetic differences between source images and their edited counterparts to acquire generalizable principles that transfer across visual domains. To support this paradigm, we construct the RED-20k dataset through a fully automated data engine that leverages VLM reasoning models and controllable editing models to generate edited variants for each source image and obtain pairwise aesthetic difference scores with chain-of-thought (CoT) reasoning.

We design a three-stage training strategy. The first stage injects aesthetic difference knowledge via supervised fine-tuning (SFT) on source-edit pairs to endow the model with rich and attributable aesthetic representations. The second stage performs a lightweight score calibration. To fully leverage the aesthetic knowledge acquired from pre-training, the third stage employs Group Relative Policy Optimization (GRPO)~\cite{shao2024deepseekmath} combined with a novel Relative Ranking Consistency Reward. This approach drives the model to produce predictions that are ordinally consistent within each source-edit group. This reinforcement learning stage enables the model to apply the causal aesthetic knowledge (\cref{fig:intro}(c)) acquired during pre-training more reliably, achieving superior generalization.

Our contributions are as follows:
\begin{itemize}
    \item We construct RED-20k, an editing-based dataset of image pairs with quantitative aesthetic differences and CoT reasoning supported by a scalable data engine with multi-model ensemble and consensus-based quality control.
    \item We propose RED-Aes, a novel IAA framework that shifts from absolute score regression to relative edit-induced difference learning using a three-stage pipeline that progresses from comparative SFT to score calibration and GRPO-based reinforcement learning to optimize the model solely via relative supervision.
    \item Extensive experiments demonstrate state-of-the-art performance on five public benchmarks where our method outperforms both specialist and generalist models by significant margins.
\end{itemize}

\section{Related Work}

\subsection{Image Aesthetic Assessment}
\label{sec:2.1}

Existing IAA datasets map single images to absolute aesthetic scores. AVA~\cite{murray2012ava} provides approximately 250k crowdsourced ratings; TAD66K~\cite{he2022rethinking} offers theme-aware annotations; AADB~\cite{kong2016photo} and PARA~\cite{yang2022personalized} enrich the space with multi-attribute labels and personalized preferences. On the modeling side, IAA has evolved from hand-crafted features~\cite{datta2006studying,marchesotti2011assessing} to deep learning. NIMA~\cite{talebi2018nima} popularized CNN-based MOS regression, followed by TANet~\cite{he2022rethinking}, MUSIQ~\cite{ke2021musiq}, and AesMamba~\cite{gao2024aesmamba} that improved architectures while retaining the same objective. More recently, VLMs have been applied to IAA. Q-Align~\cite{wu2023q} and Q-Instruct~\cite{wu2024q} leverage instruction tuning for visual scoring; AesExpert~\cite{huang2024aesexpert} generates expert-style descriptions; UNIAA~\cite{zhou2024uniaa} and ArtiMuse~\cite{cao2025artimuse} combine scoring with fine-grained textual analysis. Despite these advances, the fundamental paradigm remains unchanged: mapping individual images to absolute scores via supervised training on limited datasets.

\subsection{Controllable Image Editing}
\label{sec:2.2}

The rapid development of diffusion-based generative models has enabled precise controllable image editing, serving as the foundation of our data engine. InstructPix2Pix~\cite{brooks2023instructpix2pix} pioneered instruction-guided editing by training a conditional diffusion model on synthetic paired data. Qwen-Image-Edit~\cite{wu2025qwen} is a 20B-parameter Multi-Modal Diffusion Transformer with outstanding semantic understanding capabilities. FLUX.1 Kontext~\cite{labs2025flux} achieves high-fidelity editing through flow matching in latent space. LongCat-Image~\cite{team2025longcat} is a 6B-parameter bilingual diffusion model trained with reward-model-guided reinforcement learning. Seedream~\cite{seedream2025seedream} and Nano-Banana~\cite{google2025nanobananapro} are powerful closed-source models that deliver high-quality editing with robust performance. Using multiple architecturally distinct models ensures that aesthetic variations in our dataset reflect genuine visual changes rather than artifacts of any single pipeline.

\subsection{Post-Training with Reinforcement Learning}
\label{sec:2.3}

DeepSeek-R1~\cite{guo2025deepseek} introduced a post-training paradigm combining supervised fine-tuning with reinforcement learning via GRPO~\cite{shao2024deepseekmath}, eliciting superior logical reasoning capabilities with significantly reduced computational costs. This paradigm has been rapidly adopted in VLM-based image assessment tasks. Q-Insight~\cite{li2025q} applies GRPO for joint score regression and degradation perception. VisualQuality-R1~\cite{wu2025visualquality} trains an NR-IQA model with GRPO-based learning to rank under the Thurstone model. Aes-R1~\cite{liu2025unlocking} uses reinforcement learning to encourage more detailed aesthetic critiques. While these methods demonstrate the effectiveness of RL for assessment, they typically optimize for individual response quality or absolute score accuracy. In contrast, our approach employs GRPO with a Relative Ranking Consistency Reward that optimizes ordinal consistency within groups of aesthetically related images, directly enforcing comparative reasoning over causal aesthetic differences.

\section{Methods}
Our approach comprises two core components: (i) the RED-20k dataset, constructed through a fully automated pipeline leveraging VLMs and controllable editing models to produce relative aesthetic score differences with causal reasoning (\cref{sec:3.2}), and (ii) RED-Aes, a three-stage learning framework progressing from comparative learning to anchor calibration and GRPO-based optimization (\cref{sec:3.3}). The detailed prompts used for VLMs throughout the data construction and training pipeline are provided in the supplementary material.

\subsection{RED-20k: An Edit-based Aesthetic Difference Dataset}
\label{sec:3.2}

\begin{figure*}[t]
    \centering
    \includegraphics[width=\linewidth]{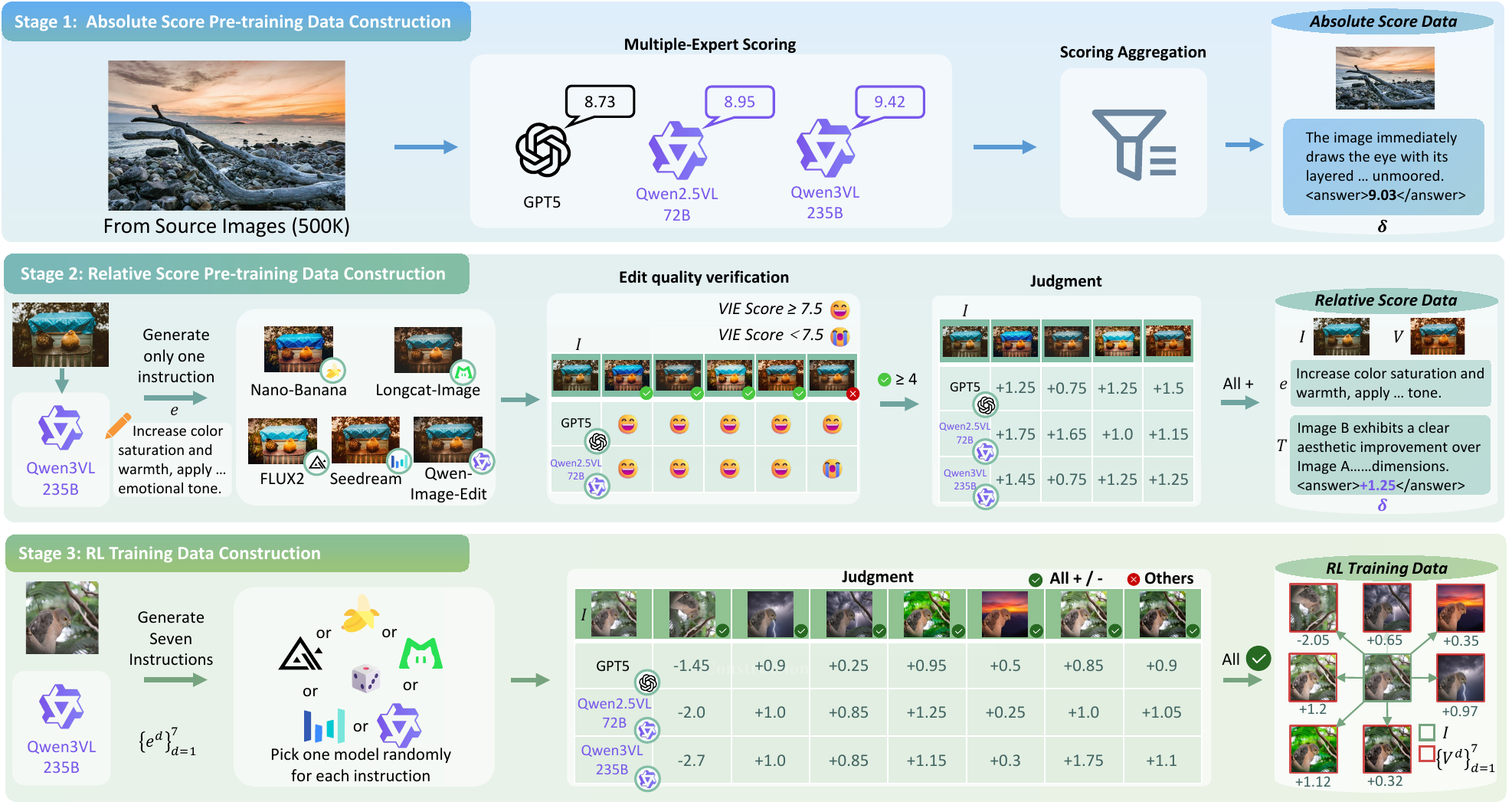}
    \caption{\textbf{Overview of the data construction pipeline.} (1) Absolute Score Data Construction, utilizing multi-expert scoring combined with score aggregation strategies to obtain absolute scores; (2) Relative Score Data Construction, utilizing diverse editing models to generate source-edited pairs, filtering via VIEScores, and employing VLM judges for consistency verification; and (3) RL Training Data Construction, which scales up to group-wise samples by generating multiple instructions and applying random editors to create comprehensive preference datasets for reinforcement learning.}
    \label{fig:data_engine}
\end{figure*}

\noindent\textbf{Source Data Collection.}
The dataset construction pipeline is illustrated in \cref{fig:data_engine}. We collect 500k images from properly licensed sources including OpenImages~\cite{krasin2017openimages}, DIV2K~\cite{agustsson2017ntire}, UHD-IQA~\cite{hosu2024aim}, Unsplash Lite~\cite{unsplashlite}, and proprietary in-house data, spanning diverse quality levels and content categories. Three VLMs (Qwen3-VL-235B, Qwen2.5-VL-72B, and GPT-5) independently score each image, and a multi-expert voting protocol produces a consensus score $s_i$ for each image $I_i$, yielding a source pool $\mathcal{O} = \{(I_i,\, s_i)\}_{i=1}^{n}$, where $n$ is the number of source images. These scores serve only as anchors for computing relative differences and generating editing instructions. The multi-expert scoring strategy utilizes MAD-based robust z-scores to detect outliers and adaptively selects a fusion method that employs weighted averaging under high consensus. Further details are provided in the supplementary material.

\noindent\textbf{Pre-training Data.}
The construction proceeds in three steps. For each source image $I$ with score $s$ from $\mathcal{O}$, Qwen3-VL-235B first analyzes its aesthetic deficiencies and generates a targeted editing instruction $e$. We apply $e$ to $I$ using five editing models $\mathcal{M} = \{m_1, \ldots, m_5\}$ (Qwen-Image-Edit, FLUX2, Longcat-Image, Nano-Banana, and Seedream-4.5), producing five edited variants $\{V_j\}_{j=1}^{5}$, where $V_j = m_j(I, e)$.

Second, we employ the VIEScore~\cite{ku2024viescore} to verify edit quality. Two VLMs (Qwen2.5-VL-72B and GPT-5) compute VIEScores, and we apply a strict threshold of 7.5 on a [0,10] scale. A source image is retained only when at least four of five models produce successful edits:
\begin{equation}
    \mathcal{V} = \{I,\, s,\, \{V_{j}^{*}\}_{j \in \mathcal{S}}\}, \quad |\mathcal{S}| \geq 4,
    \label{eq:valid_pairs}
\end{equation}
where $\mathcal{S} \subseteq \{1, \ldots, 5\}$ denotes the subset passing the quality threshold, $V_{j}^{*}$ is the edited variant produced by model $m_j$ that satisfies VIEScore $\geq 7.5$, and $\mathcal{V}$ is the retained valid source-edit record.

Finally, three judgment models $\mathcal{J} = \{J_1, J_2, J_3\}$ (Qwen3-VL-235B, Qwen2.5-VL-72B, and GPT-5) independently assess every source-edit pair in $\mathcal{V}$, predicting the direction and magnitude of aesthetic change with CoT reasoning:
\begin{equation}
    (\delta_{jk},\, T_{jk}) = J_k(I,\, V_{j}^{*},\, s), \quad j \in \mathcal{S},\; k \in \{1, 2, 3\},
    \label{eq:judge}
\end{equation}
where $\delta_{jk}$ denotes the aesthetic score difference (edited minus source) predicted by judge $J_k$ for variant $V_j^*$, and $T_{jk}$ is the corresponding CoT reasoning trace. A pair is retained only when all three judges unanimously agree on improved aesthetics. The final $\delta_{j}$ is computed via the voting protocol over $\{\delta_{jk}\}_{k=1}^3$. This strict consensus filters out noisy or ambiguous pairs. Among all valid variants, we randomly select one to form the final pre-training set $\mathcal{P}$, where each sample is a tuple $(I,\, s,\, V^{*},\, \delta,\, T,\, e)$. Here $V^{*}$ is uniformly sampled from $\{V_{j}^{*}\}_{j \in \mathcal{S}}$ and $\delta$, $T$ are the corresponding annotations.

\noindent\textbf{RL Training Data.}
We uniformly sample 1k images from $\mathcal{O}$. For each source image $I$, Qwen3-VL-235B generates seven diverse editing instructions $\{e^{d}\}_{d=1}^{7}$, each sampled from a predefined pool of 16 edit categories (\eg, style transfer, object removal, color grading). One editing model is randomly selected from $\mathcal{M}$ per instruction. The same multi-judge consensus is applied, retaining pairs only when all judges agree on the aesthetic change direction. The final RL group is:
\begin{equation}
    \mathcal{G} = \left\{(I,\, s)\right\} \cup \left\{(V^{d},\, s + \delta^{d})\right\}_{d \in \mathcal{R}},
    \label{eq:rl_group}
\end{equation}
where $V^{d}$ is the edited variant, $\delta^{d}$ the consensus difference, and $\mathcal{R} \subseteq \{1, \ldots, 7\}$ the valid subset. Each group contains eight images (one source and seven variants) forming a comparison graph for GRPO.

\noindent\textbf{Human Alignment Validation.}
We verify the reliability of labels produced by the VLMs. We conduct a human alignment study on 500 randomly sampled RED-20k pairs, with five evaluators including professional designers and lay users. The absolute aesthetic scores achieve an SRCC of 0.6843 against human annotations, while the relative source-edit differences, which serve as the core supervision signal in our framework, reach a higher SRCC of 0.7379, confirming that relative comparison is more reliably aligned with human perception than absolute scoring.

\subsection{RED-Aes: A Three-Stage Learning Framework}
\label{sec:3.3}

\begin{figure*}[t]
    \centering
    \includegraphics[width=\linewidth]{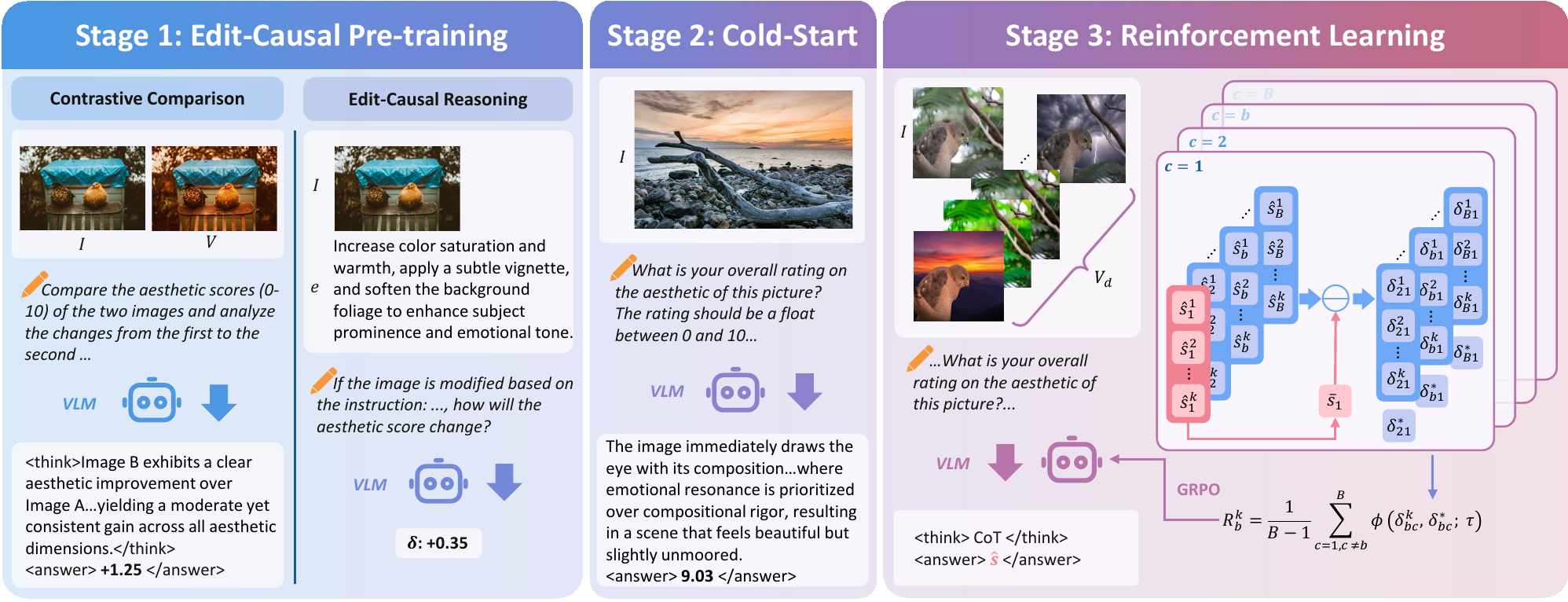}
    \caption{\textbf{The RED-Aes training pipeline.} Stage~1 injects aesthetic knowledge through two SFT targets: PT-O1 compares source-edit pairs to predict aesthetic differences with CoT reasoning, while PT-O2 predicts the aesthetic impact of editing instructions without observing the edited image. Stage~2 aligns the model to output absolute aesthetic scores using only 200 anchor samples from the source pool. Stage~3 optimizes the policy through a relative ranking consistency reward that enforces ordinal agreement with ground-truth rankings.}
    \label{fig:training_pipeline}
\end{figure*}

\subsubsection{Stage 1: Edit-Causal Pre-training.}
\label{sec:3.3.1}

The first stage injects aesthetic knowledge via supervised fine-tuning on RED-20k. As shown in \cref{fig:training_pipeline}, we focus on two optimization targets, contrastive comparison and edit-causal reasoning.

\noindent\textbf{Contrastive Comparison (PT-O1).}
Given a source-edit pair $(I, V)$, the model compares their aesthetic qualities, generates CoT reasoning describing the visual differences, and predicts the aesthetic difference $\delta$.

\noindent\textbf{Edit-Causal Reasoning (PT-O2).}
Given only the source image $I$ and the editing instruction $e$ (without the edited image), the model predicts the aesthetic difference $\delta$ that would result from the edit. This forces the model to internalize the causal relationship between visual modifications and aesthetic changes.

PT-O1 develops relative judgment from visual evidence, while PT-O2 extends this to causal prediction from textual instructions. Given the pre-training set $\mathcal{P}$, we minimize:
\begin{equation}
    \mathcal{L}_{\text{SFT}} = -\mathbb{E}_{(I,\, s,\, V^{*},\, \delta,\, T,\, e) \sim \mathcal{P}} \left[ \log p_\theta(\delta, T \mid \mathcal{X}) \right],
    \label{eq:sft_loss}
\end{equation}
where $p_\theta$ is the model distribution, $\mathcal{X} = (I, V^{*})$ for PT-O1 and $\mathcal{X} = (I, e)$ for PT-O2. Both objectives are mixed and trained jointly for 3 epochs.

\subsubsection{Stage 2: Cold-Start Calibration.}
\label{sec:3.3.2}

After pre-training, the model possesses rich aesthetic knowledge through relative comparisons but cannot yet produce absolute scores. This stage bridges the gap via format alignment on 200 samples uniformly sampled from the source pool, each annotated with a standardized absolute score and CoT reasoning. It serves as a lightweight calibration preparing for reinforcement learning. We use the same hyperparameters as Stage 1 and train for 1 epoch.

\subsubsection{Stage 3: Reinforcement Learning.}
\label{sec:3.3.3}

The final stage employs GRPO~\cite{shao2024deepseekmath} with a novel reward operating entirely on relative supervision (\cref{fig:training_pipeline}).
Consider a batch of $B$ images from the same group $\mathcal{G}$. For each image $I_b$, the policy $\pi_\theta$ samples $K$ outputs with predicted score $\hat{s}_b^{k}$. Let $\delta_{bc}^*$ denote the ground-truth score difference between $I_b$ and $I_c$. Each image serves as a comparison anchor. We compute the mean prediction $\bar{s}_c = (1/K)\sum_{k'=1}^{K} \hat{s}_c^{k'}$ for every other image $I_c$ and define the relative-ranking reward $R_b^k$ for the $k$-th output of image $I_b$ as:
\begin{equation}
    R_b^{k} = \frac{1}{B-1} \sum_{\substack{c=1,c \neq b}}^{B} \phi\!\left(\hat{s}_b^{k} - \bar{s}_c,\; \delta_{bc}^*;\; \tau\right),
    \label{eq:reward}
\end{equation}
where $\phi(\delta_{\text{pred}}, \delta_{\text{gt}}; \tau) = \tanh(|\delta_{\text{gt}}|) \cdot \sigma(\delta_{\text{pred}} \cdot \delta_{\text{gt}} / \tau)$ is a soft Kendall concordance function, $\sigma(\cdot)$ the sigmoid, and $\tau$ a temperature. The $\tanh$ term serves as a gap-aware weight emphasizing pairs with large aesthetic differences while suppressing near-ties, and the sigmoid measures soft agreement between predicted and ground-truth ordinal directions.

The advantage is computed relative to group statistics:
\begin{equation}
    A_b^{k} = \frac{R_b^{k} - \text{mean}(\{R_b^{k'}\}_{k'=1}^{K})}{\text{std}(\{R_b^{k'}\}_{k'=1}^{K})},
    \label{eq:advantage}
\end{equation}
and the policy is updated by maximizing the following GRPO objective:
\begin{equation}
    \mathcal{L}_{\text{GRPO}} = \mathbb{E}_{\substack{q \sim \mathcal{D},\; \{o_b^{k}\}_{k=1}^{K} \sim \pi_{\theta_{\text{old}}}}} \left[ \frac{1}{K} \sum_{k=1}^{K} \min\!\left( r_b^{k} A_b^{k},\; \hat{r}_b^{k} A_b^{k} \right) - \beta\, D_{\text{KL}}\!\left(\pi_\theta \| \pi_{\text{ref}}\right) \right],
    \label{eq:grpo}
\end{equation}
where $\mathcal{D}$ is the RL training distribution, $q$ is the input prompt, $o_b^{k}$ is the $k$-th sampled response for image $I_b$, $\pi_{\theta_{\text{old}}}$ is the old policy used for importance sampling, $\hat{r}_b^{k} = \text{clip}(r_b^{k},\, 1{-}\epsilon,\, 1{+}\epsilon)$, $r_b^{k} = \pi_\theta(o_b^{k} \mid q)\, /\, \pi_{\theta_{\text{old}}}(o_b^{k} \mid q)$ is the importance sampling ratio, $\epsilon$ is the clipping range, and $\beta$ controls the strength of the KL divergence penalty against the reference policy $\pi_{\text{ref}}$. This optimization encourages the model to produce scores that are ordinally consistent within each source-edit group, without requiring any absolute score supervision during the RL stage. In our implementation, we use $B = 8$, $K = 6$, and $\tau = 0.1$.

\section{Experiments}
\label{sec:exp}
We begin by demonstrating the zero-shot generalization of our model on five benchmarks where it outperforms both supervised aesthetic methods and generalist VLMs without accessing target training data. We subsequently conduct ablation studies to validate that source-consistent comparative learning effectively decouples aesthetic assessment from semantic biases. We also verify the synergy between content-oriented and quality-oriented editing instructions while determining the optimal ensemble trade-off for our data engine. We further confirm that our edit-causal learning objective and Relative Consistency Reward substantially surpass standard contrastive objectives and reasoning-based baselines. We finally investigate the impact of reinforcement learning group size and show that a contrastive context involving one source image against seven edited variants is essential for learning fine-grained rankings.

\begin{table}[t]
\centering
\caption{Computational cost of RED-20k construction.}
\label{tab:cost}
\setlength{\tabcolsep}{8pt}
\begin{tabular}{lcc}
\toprule
\textbf{Stage} & \textbf{GPU Hours (H200)} & \textbf{API Calls} \\
\midrule
Absolute Score Data Construction & 695 & 500k \\
Relative Score Data Construction & 1,100 & 500k \\
Pairwise Aesthetic Judging & 410 & 295k \\
RL Training Data Construction & 15 & 1.5k \\
\midrule
7B SFT Training & 12 & --- \\
7B RL Training & 560 & --- \\
\bottomrule
\end{tabular}
\end{table}

\subsection{Experiment Settings}
\label{sec:4.1}
\noindent\textbf{Datasets and Metrics.} We evaluate our method on five public benchmarks including TAD66K~\cite{he2022rethinking}, AVA~\cite{murray2012ava}, FLICKR-AES~\cite{ren2017personalized}, PARA~\cite{yang2022personalized}, and AADB~\cite{kong2016photo}. All experiments operate under a cross-domain setting where the model is trained exclusively on our proposed dataset. We normalize the ground-truth Mean Opinion Scores from their original ranges of 1-10 for TAD66K and AVA or 1-5 for others to the [0, 1] interval. We utilize Pearson Linear Correlation Coefficient (PLCC)~\cite{benesty2009pearson} to measure prediction accuracy and Spearman Rank-Order Correlation Coefficient (SRCC)~\cite{sedgwick2014spearman} to assess monotonicity.

\noindent\textbf{Evaluation Settings.} We report results on all five datasets for main comparisons and conduct ablation studies on the larger TAD66K, AVA, and FLICKR-AES benchmarks to ensure statistical robustness. All models function in a zero-shot manner without fine-tuning on target datasets. We run open-source models using recommended inference settings for baselines that lack public results. The prompts used for all evaluated models are provided in the supplementary material to ensure reproducibility.

\noindent\textbf{Implementation Details.} We implement the RED-Aes using Qwen2.5-VL-7B~\cite{Bai2025Qwen25VLTR}, Qwen2.5-VL-3B~\cite{Bai2025Qwen25VLTR}, and Qwen3-VL-2B~\cite{bai2025qwen3} backbones with frozen vision encoders and a fine-tuned language component. The training pipeline utilizes the AdamW optimizer with $\beta_1=0.9$ and $\beta_2=0.999$ plus weight decay 0.01 and resizes images to $448 \times 448$. We first conduct supervised fine-tuning on our synthesized dataset for 3 epochs with a batch size of 16 and a learning rate of 1e-5. We subsequently execute a format alignment stage for 1 epoch using 200 samples to standardize scalar output. We finally employ reinforcement learning via GRPO to optimize ranking consistency by constructing training groups of 8 images and generating 6 sampled outputs per prompt.

\noindent\textbf{Computational Cost.} During dataset construction, the open-source Qwen-series models are deployed on H200 GPUs for local inference, while GPT-5 is accessed via API calls. These large-scale models are used exclusively for offline dataset construction. \cref{tab:cost} details the computational cost of each construction stage and model training, including GPU hours for locally deployed models and the number of API calls. Furthermore, the RL stage operates on only 1k source images and their corresponding edited variants, and the entire annotation and editing pipeline is sample-independent and highly parallelizable.

\begin{table}[t]
\centering
\caption{Quantitative evaluation on multiple public benchmarks. $^{\dagger}$ Results are reproduced by us using the official open-source models as the original papers did not report performance on these specific datasets.}
\setlength{\tabcolsep}{4pt}
\renewcommand\arraystretch{0.95}
\resizebox{1.0\linewidth}{!}{
\begin{tabular}{l cc cc cc cc cc | cc}
\toprule
\multirow{2}{*}{\textbf{Model}} 
  & \multicolumn{2}{c}{\textbf{TAD66K}} & \multicolumn{2}{c}{\textbf{AVA}} & \multicolumn{2}{c}{\textbf{FLICKR}} & \multicolumn{2}{c}{\textbf{PARA}} & \multicolumn{2}{c|}{\textbf{AADB}} & \multicolumn{2}{c}{\textbf{Average}} \\
\cmidrule(lr){2-3} \cmidrule(lr){4-5} \cmidrule(lr){6-7} \cmidrule(lr){8-9} \cmidrule(lr){10-11} \cmidrule(lr){12-13}
& PLCC & SRCC & PLCC & SRCC & PLCC & SRCC & PLCC & SRCC & PLCC & SRCC & PLCC & SRCC\\
\midrule
\multicolumn{13}{l}{\textit{Vanilla VLM}} \\
Qwen2.5-VL-7B~\cite{Bai2025Qwen25VLTR}     & 0.2282 & 0.2242 & 0.3518 & 0.3684 & 0.5179 & 0.5696 & 0.5966 & 0.6252 & 0.4480 & 0.5076 & 0.4285 & 0.4589 \\
Qwen2.5-VL-32B~\cite{Bai2025Qwen25VLTR}     & 0.2300 & 0.2715 & 0.3893 & 0.4170 & 0.5800 & 0.6527 & 0.7176 & 0.7118 & 0.5060 & 0.5138 & 0.4846 & 0.5134   \\
Qwen3-VL-8B~\cite{bai2025qwen3}     & 0.2642 & 0.2464 & 0.4663 &  0.4599 & 0.6258 & 0.6190 & 0.6050 &0.6127 & 0.6934 & 0.6460 & 0.5309 & 0.5168 \\
Qwen3-VL-32B~\cite{bai2025qwen3}     &0.3042 &0.2816 & 0.5057& 0.4955& 0.6651 & 0.6545& 0.7176 & 0.7118 &  0.7147 &0.6782 & 0.5815 & 0.5643   \\
Seed1.5-VL-thinking~\cite{guo2025seed1}                 & 0.2908 & 0.2974 & 0.4533 & 0.4871 & 0.5837 & 0.6374 & 0.6134 & 0.6364 & 0.5279 & 0.5577 & 0.4938 & 0.5232 \\
GPT-4o~\cite{hurst2024gpt}                   & 0.2686 & 0.2979 & 0.4687 & 0.4939 & 0.5469      &  0.5507     & 0.6916 & 0.6719 & 0.5209 & 0.5314 & 0.4993 & 0.5092 \\
GPT-5~\cite{singh2025openai}                & 0.2973 & 0.3347 & 0.4852 & 0.5447 & 0.6218 & 0.6260 & 0.7433 & 0.7352 & 0.6001 & 0.5989 & 0.5495 & 0.5679 \\
Multi-Expert Voting              & 0.4925 & 0.5031 & 0.6654 & 0.6743 & 0.7175 & 0.7219 & 0.8397 & 0.7957 & 0.7124 & 0.6765 & 0.6855 & 0.6743 \\
\midrule
\multicolumn{13}{l}{\textit{Deep-Learning Based}} \\
NIMA~\cite{talebi2018nima}                     & 0.3885 & 0.3654 & 0.6120 & 0.6361& 0.5130 & 0.4796 & 0.5868 & 0.5709 & 0.3886 & 0.3904 & 0.4978 & 0.4885 \\
\midrule
\multicolumn{13}{l}{\textit{VLM Based}} \\
ArtiMuse-8B~\cite{cao2025artimuse}                  & 0.2320 & 0.2300 & 0.3850 & 0.3970 & 0.3340 & 0.3490 & 0.6057$\dagger$ & 0.5624$\dagger$ & 0.4836$\dagger$ & 0.5130$\dagger$ & 0.4081 & 0.4103 \\
Aes-R1-7B~\cite{liu2025unlocking} & 0.4513& 0.4248 & 0.6702  & 0.6619  & 0.7243 & 0.6973& 0.7842 & 0.7666& 0.5386 & 0.5423 & 0.6337& 0.6186 \\
UniPercept-8B~\cite{cao2025unipercept}  & 0.3460& 0.3360 & 0.5770  & 0.5890  & 0.6810 & 0.6880 & 0.7006$\dagger$ & 0.5869$\dagger$ & 0.4285$\dagger$ & 0.4237$\dagger$ & 0.5466 & 0.5247 \\
\midrule
RED-Aes-7B (Ours) & \textbf{0.5406} & \textbf{0.5322} & \textbf{0.7713}  & \textbf{0.7700}  & \underline{0.7524} & \underline{0.7252}& \textbf{0.8846} & \textbf{0.8591} & \textbf{0.7722} & \textbf{0.7746} & \textbf{0.7442} & \textbf{0.7322} \\
RED-Aes-3B (Ours) & \underline{0.5275}&\underline{0.5297}& \underline{0.7518} &\underline{0.7514}  & 0.7292 & 0.7089 & 0.8547 & \underline{0.8413} & 0.7365 & 0.7509 & \underline{0.7199} & \underline{0.7164}\\
RED-Aes-2B (Ours) & 0.4784& 0.4778 & 0.7235 & 0.7148  & \textbf{0.7565} & \textbf{0.7401} & \underline{0.8616} & 0.8360 & \underline{0.7426} & \underline{0.7523} & 0.7125 & 0.7042\\
     \bottomrule 
\end{tabular}
}
\label{tab:multidataset}
\end{table}

\subsection{Comparison with State-of-the-Art}
\label{sec:4.3}
We evaluate our method on five benchmarks including TAD66K, AVA, FLICKR, PARA, and AADB under a rigorous zero-shot cross-dataset setting. Our model is trained exclusively on the proposed dataset. As shown in \cref{tab:multidataset}, our model based on Qwen2.5-VL-7B achieves state-of-the-art performance with an average PLCC of 0.7442 and SRCC of 0.7322. It substantially surpasses UniPercept~\cite{cao2025unipercept} and GPT-5, which represent the leading generalist aesthetic models. Furthermore, it outperforms Aes-R1 which serves as the strongest specialist model trained directly on in-domain data. This performance is particularly notable considering that our approach generalizes from synthesized comparative pairs, whereas baselines rely heavily on fitting target-domain absolute scores. Remarkably, even our lightweight variant based on Qwen3-VL-2B outperforms all baselines to demonstrate exceptional efficiency. We also compare against a zero-shot multi-expert voting baseline that directly ensembles scores from the three VLMs used in our data construction (Qwen3-VL-235B, Qwen2.5-VL-72B, and GPT-5). This ensemble achieves an average SRCC of only 0.6743 with prohibitive deployment cost, falling below even our 2B model, confirming the effectiveness of our training paradigm in distilling aesthetic reasoning capabilities into compact models.

We attribute these gains to the comparative supervision provided by our rigorously constructed dataset. First, the relative-score pre-training stage injects comparative reasoning capabilities that allow the model to learn universal aesthetic gradients rather than overfitting to absolute scalar values. Second, the reinforcement learning stage utilizes our relative ranking consistency reward to enforce ordinal relationships among images. This design aligns with the inherently comparative nature of human perception. The consistent improvements observed across Qwen2.5-VL and Qwen3-VL backbones further validate that our method is architecture-agnostic and generalizes effectively across different VLM families.

\begin{table}[t]
    \centering
    \caption{Ablation Study on Dataset Quality and Training Consistency. We progressively evaluate the contribution of our collected dataset (Rows 1-2) and the impact of the group-wise consistent training strategy (Rows 3-4) across three benchmarks.}
    \label{tab:comprehensive_ablation}
    \resizebox{0.9\linewidth}{!}
     {
    \begin{tabular}{l|cccccc|cc}
        \toprule
        \multirow{2}{*}{\textbf{Configuration}} & \multicolumn{2}{c}{\textbf{AVA}} & \multicolumn{2}{c}{\textbf{FLICKR}} & \multicolumn{2}{c|}{\textbf{TAD66K}} & \multicolumn{2}{c}{\textbf{Average}} \\
        
         \cmidrule(lr){2-3} \cmidrule(lr){4-5} \cmidrule(lr){6-7} \cmidrule(lr){8-9}
        & PLCC & SRCC& PLCC& SRCC& PLCC & SRCC & PLCC & SRCC\\
        \midrule
         Aes-R1 \textit{w}/ AesCoT-15k & 0.6702 & 0.6619 & 0.7243 & 0.6973&0.4513 & 0.4248& 0.6153 & 0.5947 \\
        Aes-R1 \textit{w}/ RED-20k  & 0.7506 & 0.7381 & 0.7270&0.6983 &0.5107 & 0.5032 &0.6628 & 0.6465 \\
        \midrule
        Ours (Shuffled Batch) &0.7564 &0.7423 & 0.7281&0.7038 & 0.5184 & 0.5134&0.6676& 0.6532 \\
        Ours (Consistent Batch)& \textbf{0.7713} & \textbf{0.7700} & \textbf{0.7524} & \textbf{0.7252} & \textbf{0.5406} & \textbf{0.5322} & \textbf{0.6881} & \textbf{0.6758} \\
        \bottomrule
    \end{tabular}
    }
\end{table}

\subsection{Efficacy of Dataset and Group-wise Consistency}
\label{sec:4.4}
To disentangle the contributions of our dataset from our training paradigm, we conduct a progressive ablation study as presented in \cref{tab:comprehensive_ablation}.

\noindent\textbf{Dataset Superiority.}
We first validate the quality of our collected dataset by replicating the training pipeline of the Aes-R1 baseline but substituting their original data with our proposed dataset. As indicated in the comparison between the first two rows, merely transitioning to our dataset yields a substantial performance gain where the average SRCC increases from 0.5947 to 0.6465. This significant margin confirms that our dataset provides supervision signals of superior quality and robustness compared to prior benchmarks. 

\noindent\textbf{Importance of Consistent Context.}
Next, we investigate the impact of group-wise consistency. We compare our strategy against a baseline that shuffles images within the batch while preserving the pairing between each image and its original editing instruction, thereby forcing comparisons between unrelated content. Results demonstrate that the shuffled setting underperforms our consistent approach with an average SRCC of 0.6532 versus 0.6758. We attribute this gap to content-dependent biases, as comparing disparate images leads the model to learn spurious correlations such as category preferences instead of genuine aesthetic improvements. By contrast, our strategy isolates aesthetic changes from the content so that the model focuses purely on the relative quality of edits.

\subsection{Impact of Edit Types}
\label{sec:4.5}
We categorize editing instructions into content-oriented tasks (e.g., inpainting) and quality-oriented tasks (e.g., color enhancements). \cref{tab:ablation_edit_types} presents the evaluation of models trained on various combinations of these types.

\noindent\textbf{Synergy between Content and Quality.}
We observe that neither content-only nor quality-only training yields optimal performance. The full model integrates both types during pre-training and reinforcement learning stages and consistently outperforms the ablated variants. For instance, it increases the average SRCC from 0.6421 in the quality-only setting to 0.6758. This confirms that robust assessment benefits from understanding both low-level perceptual attributes and high-level semantic composition.

\noindent\textbf{Necessity of Diverse RL Supervision.}
We further compare the full model against the variant using mixed pre-training with quality-only reinforcement learning. Excluding content edits from the alignment stage leads to a distinct performance drop across all datasets. Specifically, the SRCC on AVA falls from 0.7700 to 0.7379. These results demonstrate that reinforcing ranking consistency across diverse edit types is crucial for generalization.

\begin{table}[t]
    \centering
    \caption{Ablation study on editing instruction types. We report PLCC and SRCC across three benchmarks. ``Cont.'' and ``Qual.'' denote content-oriented and quality-oriented editing instructions, respectively. ``PT'' refers to Pre-training and ``RL'' to Reinforcement Learning. ``Mixed'' implies using both instruction types.}
    \label{tab:ablation_edit_types}
    \resizebox{0.95\linewidth}{!}
    {
    \begin{tabular}{l|cccccc|cc}
        \toprule
        \multirow{2}{*}{\textbf{Method Setting}} & \multicolumn{2}{c}{\textbf{AVA}} & \multicolumn{2}{c}{\textbf{FLICKR}} & \multicolumn{2}{c|}{\textbf{TAD66K}} & \multicolumn{2}{c}{\textbf{Average}} \\
        
        \cmidrule(lr){2-3} \cmidrule(lr){4-5} \cmidrule(lr){6-7} \cmidrule(lr){8-9}
         & PLCC& SRCC& PLCC & SRCC & PLCC& SRCC& PLCC& SRCC \\
        \midrule

        Cont. Only PT \& RL    & 0.7050 & 0.6939 & 0.7234 & 0.6987 & 0.5144 & 0.5020 & 0.6476 & 0.6315 \\
        Qual. Only PT \& RL   & 0.7152 & 0.7130 & 0.7221 & 0.6944 & 0.5238 & 0.5188 & 0.6537 & 0.6421 \\
        Mixed PT + Qual. RL      & 0.7426 & 0.7379 & 0.7450 & 0.7047 & 0.5317 & 0.5202 & 0.6731 & 0.6543 \\
        \midrule

        Mixed PT \& RL & \textbf{0.7713} & \textbf{0.7700} & \textbf{0.7524} & \textbf{0.7252} & \textbf{0.5406} & \textbf{0.5322} & \textbf{0.6881} & \textbf{0.6758} \\
        \bottomrule
    \end{tabular}
    }
\end{table}

\subsection{Impact of Ensemble Size}
\label{sec:ablation_ensemble}
\noindent\textbf{Number of Editing Models ($N_E$).}
We conducted a pilot study on 5k raw images to determine the optimal configuration. We first investigated editing diversity by varying $N_E$ from 1 to 7. \cref{tab:ablation_model_num} shows that increasing $N_E$ to 5 yields substantial PLCC and SRCC gains and effectively captures diverse aesthetic causality. A further increase to 7 offers only marginal improvements. This indicates that five editing models sufficiently cover the variation space.

\noindent\textbf{Number of Judgment Models ($N_J$).}
We subsequently analyzed the consensus mechanism with $N_E$ fixed at 5. Transitioning from 1 to 3 judgment models significantly enhances performance by filtering label noise. Increasing $N_J$ to 5 yields negligible average gains and even slight degradation on generalization benchmarks. The configuration with 5 editing and 5 judgment models imposes a higher computational burden without meaningful benefits. We therefore adopted the strategy of 5 editing and 3 judgment models as the optimal trade-off.

\begin{table}[t]
    \centering
    \caption{Ablation study on the ensemble size of editing and judgment models. We denote $N_E$ as the number of editing models and $N_J$ as the number of aesthetic judgment models. The chosen setting ($5+3$) achieves the best trade-off between performance and computational cost.}
    \label{tab:ablation_model_num}
    \resizebox{0.9\linewidth}{!}
    {
    \begin{tabular}{cc|cccccc|cc}
        \toprule
        \multicolumn{2}{c|}{\textbf{Number Setting}} & \multicolumn{2}{c}{\textbf{AVA}} & \multicolumn{2}{c}{\textbf{FLICKR}} & \multicolumn{2}{c|}{\textbf{TAD66K}} & \multicolumn{2}{c}{\textbf{Average}} \\
         \cmidrule(lr){1-2} \cmidrule(lr){3-4} \cmidrule(lr){5-6} \cmidrule(lr){7-8} \cmidrule(lr){9-10}
        \hspace{10pt}$N_E$ & \hspace{10pt}$N_J$ & PLCC & SRCC& PLCC & SRCC & PLCC & SRCC & PLCC& SRCC \\
        \midrule
        \hspace{10pt}1 & \hspace{10pt}1 &0.6940 & 0.6834 & 0.6327 & 0.6106 & 0.4917 & 0.4876&0.6061 & 0.5939 \\
        \hspace{10pt}3 & \hspace{10pt}1 & 0.7379 & 0.7165 &0.6800 & 0.6477 & 0.5159 & 0.4969 & 0.6446& 0.6204 \\
        \hspace{10pt}5 & \hspace{10pt}1 & 0.7442 & 0.7309 & 0.7055 & 0.6667 &0.5224 & 0.5029 & 0.6574 & 0.6335\\
        \hspace{10pt}7 & \hspace{10pt}1 & 0.7459 & 0.7346 & 0.7089 & 0.6630 &0.5214 & 0.5073 & 0.6587& 0.6350 \\
        \midrule
        \hspace{10pt}5 & \hspace{10pt}3 & 0.7713 &0.7700 & 0.7524 & \textbf{0.7252}&  \textbf{0.5406} & 0.5322 & 0.6881&0.6758\\
        \hspace{10pt}5 & \hspace{10pt}5 & \textbf{0.7735}& \textbf{0.7722} &  \textbf{0.7531} & 0.7234 &0.5383 &  \textbf{0.5330} &  \textbf{0.6883} &  \textbf{0.6762}\\
        \bottomrule
    \end{tabular}
    }
\end{table}

\subsection{Component Analysis}
\label{sec:ablation_component}
\noindent\textbf{Effectiveness of Pre-training.}
We first examine pre-training objectives by training the model on the full dataset with supervised fine-tuning (SFT). Our dataset provides absolute aesthetic scores for all images, derived from expert ratings and relative comparisons, enabling standard SFT on the entire data. As shown in Phase 1 of \cref{tab:ablation_component}, the SFT only baseline achieves an average SRCC of only 0.5577. Adding our contrastive objective (PT-O1) raises SRCC to 0.5809. Further incorporating edit causal learning (PT-O2) boosts it to 0.6218. This improvement validates that combining static representation learning with dynamic causal modeling outperforms vanilla SFT.

\noindent\textbf{Superiority of Our RL.}
In Phase 2, we evaluate RL strategies starting from a cold-start base. This base is constructed by first pre-training the model on the full dataset with PT-O1 \& PT-O2 to learn comparative aesthetic knowledge, followed by a cold-start supervised fine-tuning (SFT) on only 200 samples with absolute score annotations to enable calibrated absolute score output (see \cref{sec:4.1}). Due to limited data, its SRCC drops to 0.5907, which is expected and provides a controlled starting point for RL comparison. On this base, Aes-R1 improves SRCC to 0.6465. Our RL method further pushes it to 0.6758, surpassing both Aes-R1 and the fully-supervised Phase 1 model (0.6218). This demonstrates that our relative-aesthetic reward aligns better with human preferences than general-purpose reasoning RL.

\begin{table}[t]
    \centering
    \caption{Component ablation study. We evaluate the incremental effectiveness of pre-training objectives (Phase 1) and reinforcement learning strategies (Phase 2) on three aesthetic benchmarks.}
    \label{tab:ablation_component}
     \resizebox{0.9\linewidth}{!}
     {
    \begin{tabular}{lcccccc|cc}
        \toprule
        \multirow{2}{*}{\textbf{Method Setting}} & \multicolumn{2}{c}{\textbf{AVA}} & \multicolumn{2}{c}{\textbf{FLICKR}} & \multicolumn{2}{c|}{\textbf{TAD66K}} & \multicolumn{2}{c}{\textbf{Average}} \\
        \cmidrule(lr){2-3} \cmidrule(lr){4-5} \cmidrule(lr){6-7} \cmidrule(lr){8-9}
         & PLCC& SRCC & PLCC& SRCC& PLCC& SRCC & PLCC & SRCC \\
        \midrule
        \multicolumn{9}{l}{\textit{Phase 1: Pre-training Strategies (Full SFT)}} \\
        SFT only & 0.6612 & 0.6332 & 0.6618 & 0.6390 &0.4344 & 0.4009 & 0.5858 &0.5577 \\
        SFT \textit{w}/ PT-O1 &0.6849 & 0.6685 &0.6878&0.6456 & 0.4550& 0.4287& 0.6092 & 0.5809 \\
        SFT \textit{w}/ PT-O1 \& PT-O2 & 0.7398 & 0.7129 &0.7159 & 0.6733 &0.4906 & 0.4791 & 0.6488 &0.6218 \\
        \midrule
        \multicolumn{9}{l}{\textit{Phase 2: RL Strategies (Few-shot Cold Start)}} \\
        Cold Start Base & 0.7023 & 0.6790&0.6911 &0.6510& 0.4719&0.4421&0.6218 & 0.5907 \\
        Cold Start \textit{w}/ Aes-R1 &0.7506& 0.7381 &0.7270 &0.6983& 0.5107&0.5032& 0.6628 & 0.6465 \\
        Cold Start \textit{w}/ Our RL & \textbf{0.7713} & \textbf{0.7700} & \textbf{0.7524} & \textbf{0.7252} & \textbf{0.5406} & \textbf{0.5322} & \textbf{0.6881} & \textbf{0.6758} \\
        \bottomrule
    \end{tabular}
    }
\end{table}

\begin{table}[t]
    \centering
    \caption{\textbf{Ablation on the Group Size (Batch Size) in RL.} In our framework, a ``batch'' consists of 1 source image and $B-1$ edited variants. We study how the number of edited candidates per image affects the optimization. $B=8$ (1 source + 7 edits) provides the most effective contrastive context.}
    \label{tab:ablation_batch_size}
    \resizebox{0.9\linewidth}{!}
     {
    \begin{tabular}{c|cccccc|cc}
        \toprule
        \multirow{2}{*}{\textbf{Batch Size ($B$)}} & \multicolumn{2}{c}{\textbf{AVA}} & \multicolumn{2}{c}{\textbf{FLICKR}} & \multicolumn{2}{c|}{\textbf{TAD66K}} & \multicolumn{2}{c}{\textbf{Average}} \\
        & \textbf{PLCC} & \textbf{SRCC} & \textbf{PLCC} & \textbf{SRCC} & \textbf{PLCC} & \textbf{SRCC} & \textbf{PLCC} & \textbf{SRCC} \\
        \midrule
        4 (1+3) & 0.7579 & 0.7483 &0.7325 & 0.7110 &0.5205 &0.5160 & 0.6703& 0.6584 \\
        6 (1+5) & 0.7676 &0.7599 & 0.7434 &0.7240 & 0.5347 &0.5246 &0.6819 & 0.6695 \\
        \textbf{8 (1+7)} & \textbf{0.7713} & \textbf{0.7700} & 0.7524 & 0.7252 & \textbf{0.5406} & \textbf{0.5322} & \textbf{0.6881} & \textbf{0.6758} \\
        10 (1+9) & 0.7701&0.7688 & \textbf{0.7576} & \textbf{0.7257} &0.5361 &0.5267&0.6879 &0.6737 \\
        \bottomrule
    \end{tabular}
    }
\end{table}

\subsection{Impact of RL Group Size}
\label{sec:ablation_batch}
\noindent\textbf{Effect of Contrastive Diversity.}
We investigate the impact of the group size $B$ (comprising one source image and $B-1$ edited variants) on the optimization process. As shown in \cref{tab:ablation_batch_size}, a small group size ($B=4$) yields suboptimal performance, achieving an average SRCC of only 0.6584, because the model observes limited aesthetic variations, restricting its ability to distinguish subtle improvements from noise.
Increasing $B$ to 8 brings significant gains, boosting the average SRCC to 0.6758. This richer contrastive context enables the model to compare the source against a diverse set of superior and inferior variants, facilitating the learning of fine-grained ranking signals.
However, further expanding the size to $B=10$ leads to performance saturation and a slight decline in the average SRCC. This suggests that 7 edited variants are sufficient to estimate the local aesthetic gradient, while adding excessive candidates may introduce redundancy or increase optimization difficulty. We therefore adopt $B=8$ as the default setting.

\begin{figure*}[t]
    \centering
    \includegraphics[width=\linewidth]{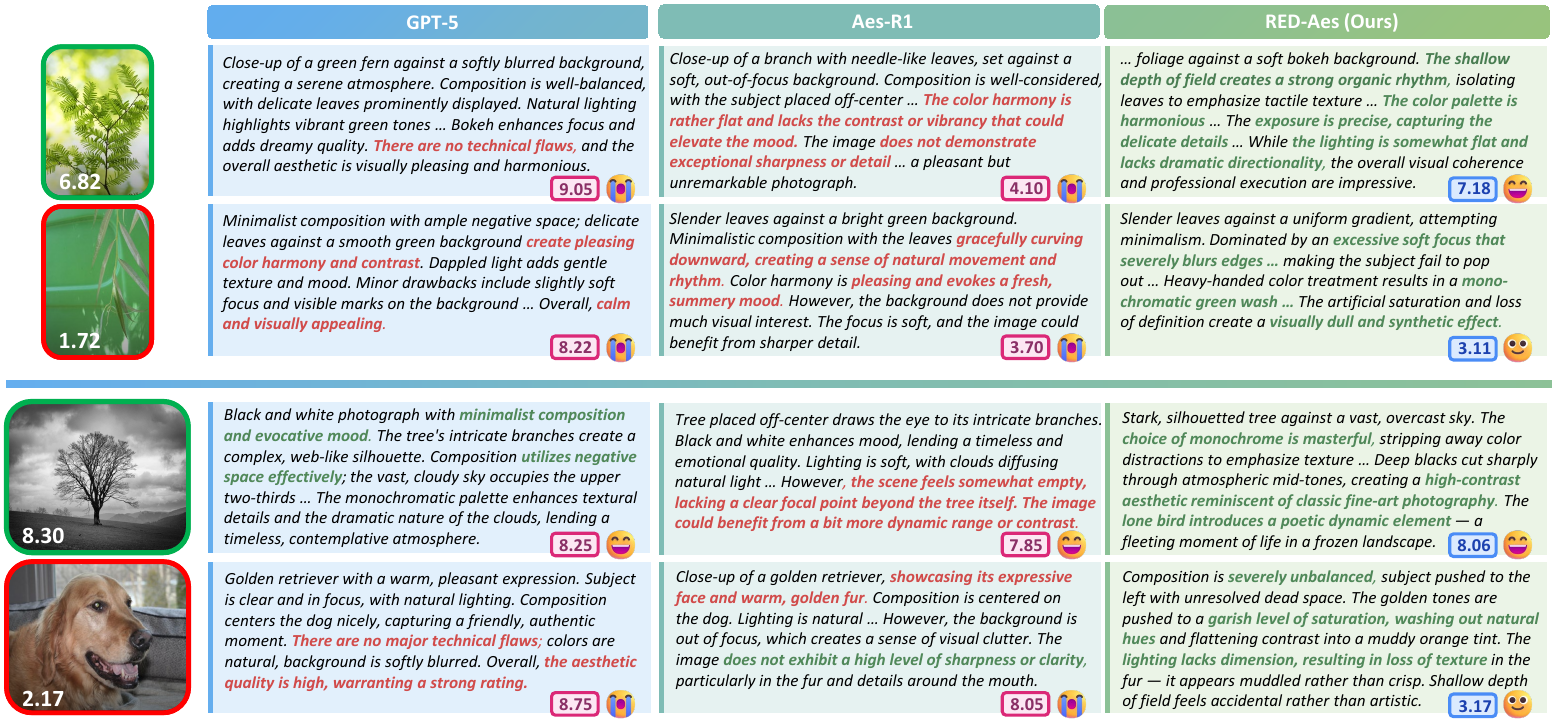}
    \caption{Qualitative comparison of RED-Aes against baselines. Each column displays the input images with Ground Truth score, and the predicted scores from each model, along with the CoT reasoning.}
    \label{fig:case_study}
\end{figure*}

\subsection{Case Study}

As illustrated in \cref{fig:case_study}, both generic VLMs and absolute-score baselines struggle to decouple semantic content from aesthetic quality.
In the second row, GPT-5 and Aes-R1 assign inflated scores of 8.22 and 3.70 to a heavily blurred image (GT 1.72), failing to penalize degradation. RED-Aes predicts 3.11 by explicitly critiquing the ``excessive soft focus'' and ``synthetic effect''.
In the fourth row, despite severe artifacts (GT 2.17), Aes-R1 and GPT-5 still predict 8.05 and 8.75, misled by semantic content such as the dog's ``expressive face''. RED-Aes correctly identifies the ``garish saturation'' and ``muddled texture,'' yielding 3.17. These results confirm that our relative edit-induced learning effectively diagnoses the specific visual factors that influence aesthetic scores.

\section{Conclusion}

We propose RED-Aes, a framework that shifts IAA from absolute score regression to relative edit-induced difference learning. By constructing source-edit pairs via controllable editing models, our approach enables the model to learn causal factors underlying aesthetic perception. The accompanying RED-20k dataset provides the first resource of editing-based aesthetic pairs with quantitative differences and CoT reasoning, built through a fully automated pipeline without human annotation. To progressively build aesthetic reasoning capabilities, we employ a three-stage training strategy comprising edit-causal pre-training, format alignment, and GRPO with a relative ranking consistency reward. Experiments on five benchmarks under zero-shot cross-domain evaluation demonstrate state-of-the-art performance, with even a lightweight 2B variant surpassing all existing baselines.

%
%
\bibliographystyle{splncs04}
\bibliography{main}


\appendix

\section{Multi-expert Voting Protocol}
\label{sec:supp_voting}

We employ a multi-expert voting protocol to produce consensus aesthetic scores for the source pool $\mathcal{O}$. Given $N$ expert models, each image receives $N$ independent scores $\{s_1, s_2, \ldots, s_N\}$. The protocol proceeds in three steps: outlier detection, consensus-level classification, and adaptive score fusion.

\noindent\textbf{Step 1: Outlier Detection via Robust Z-Scores.}
We adopt the Median Absolute Deviation (MAD) to identify anomalous scores that deviate significantly from the expert consensus. For each image, we compute the median score $\tilde{s} = \text{median}(\{s_n\}_{n=1}^{N})$ and the MAD:
\begin{equation}
    \text{MAD} = \text{median}\left(\left\{|s_n - \tilde{s}|\right\}_{n=1}^{N}\right).
\end{equation}
The modified z-score~\cite{iglewicz1993volume} for each expert $n$ is then:
\begin{equation}
    z_n = \frac{0.6745 \cdot (s_n - \tilde{s})}{\text{MAD}},
\end{equation}
where the constant $0.6745$ ensures consistency with the standard normal distribution. An expert score is flagged as an outlier if $|z_n| > \lambda$, where $\lambda = 1.5$ is the outlier threshold. This MAD-based approach is more robust than standard deviation-based methods~\cite{leys2013detecting}, as it is resistant to the influence of extreme values.

\noindent\textbf{Step 2: Consensus-Level Classification.}
We classify the agreement among experts into three levels based on the standard deviation $\sigma$ of the $N$ scores:
\begin{equation}
    \text{Consensus Level} =
    \begin{cases}
        \textit{High},    & \text{if } \sigma < \tau_h, \\
        \textit{Medium},  & \text{if } \tau_h \leq \sigma < \tau_m, \\
        \textit{Low},     & \text{if } \sigma \geq \tau_m,
    \end{cases}
\end{equation}
where $\tau_h = 0.5$ and $\tau_m = 1.0$. The consensus level reflects the degree of inter-expert agreement and directly determines the fusion strategy applied in the next step.

\noindent\textbf{Step 3: Adaptive Score Fusion.}
We design a strategy matrix indexed by the consensus level and the outlier situation (no outliers, one outlier, or multiple outliers), as shown in \cref{tab:strategy_matrix}. Each cell specifies a fusion method selected from the following:

\begin{itemize}
    \item \textbf{Weighted Average}: $\hat{s} = \sum_{n} w_n s_n / \sum_{n} w_n$, where $\mathbf{w}$ is a consensus-level-dependent weight vector.
    \item \textbf{Normal Average}: The arithmetic mean of non-outlier scores.
    \item \textbf{Weighted Median}: The median computed with consensus-level-dependent weights, which reduces sensitivity to skewed distributions.
    \item \textbf{Median}: The standard median of all scores.
    \item \textbf{Conservative Average}: A blend of $70\%$ weighted average and $30\%$ median, balancing central tendency with robustness.
    \item \textbf{Robust Median}: A Tukey's biweight estimator~\cite{beers1990measures} that down-weights scores far from the median, providing strong resistance to outliers.
\end{itemize}

In our experiments, we employ $N=3$ expert models (Qwen3-VL-235B~\cite{bai2025qwen3}, Qwen2.5-VL-72B~\cite{Bai2025Qwen25VLTR}, and GPT-5~\cite{singh2025openai}). For both the Weighted Average and Weighted Median fusion methods, the weight vector $\mathbf{w} = (w_1, w_2, w_3)$ is set uniformly as $w_1 : w_2 : w_3 = 1 : 1 : 1$ across all consensus levels, reflecting the assumption that each expert contributes equally reliable aesthetic judgments. This uniform weighting avoids introducing additional hyperparameters and ensures that the adaptive behavior of the protocol is driven entirely by the consensus-level classification and outlier detection, rather than by pre-assigned expert preferences.

\begin{table}[t]
    \centering
    \caption{Adaptive fusion strategy matrix. Each cell specifies the fusion method selected based on the consensus level and the number of detected outliers.}
    \label{tab:strategy_matrix}
    \resizebox{\linewidth}{!}{
    \begin{tabular}{lccc}
        \toprule
        \textbf{Consensus Level} & \textbf{No Outliers} & \textbf{One Outlier} & \textbf{Multiple Outliers} \\
        \midrule
        High ($\sigma < 0.5$)    & Weighted Average     & Normal Average       & Weighted Median \\
        Medium ($0.5 \leq \sigma < 1.0$) & Weighted Average & Normal Average  & Median \\
        Low ($\sigma \geq 1.0$)  & Conservative Average & Median               & Robust Median \\
        \bottomrule
    \end{tabular}
    }
\end{table}

The underlying principle is intuitive: when experts exhibit high agreement, we trust the weighted average to preserve fine-grained score distinctions; as disagreement increases, we progressively shift toward median-based estimators that are more resistant to noise. When outliers are present under high consensus, we exclude them and average the remaining scores; under low consensus with multiple outliers, we resort to the robust median to ensure stability.

This adaptive protocol ensures that the consensus scores in $\mathcal{O}$ are both accurate under agreement and stable under disagreement, providing reliable anchors for the subsequent relative difference computation.

\section{Prompt Template}
\label{sec:supp_prompt}

This section provides the complete prompt templates used throughout our pipeline, as referenced in the main paper. We organize them into three groups according to their roles: (1)~\textbf{Data Construction Prompts} (\ie, prompts for aesthetic scoring and editing instruction generation, VIEScore-based edit quality verification, pairwise aesthetic comparison, and diverse editing instruction generation for the RL phase), which drive the automated RED-20k data engine; (2)~\textbf{Training Stage Prompts}, which define the question templates for constructing Q\&A pairs across the three training stages (Stage~1 SFT with PT-O1 and PT-O2, Stage~2 cold-start calibration, and Stage~3 GRPO); and (3)~\textbf{Inference Prompt}, which specifies the evaluation prompt used by RED-Aes at test time. For baseline models, we adopt the inference prompts provided in their respective original papers or open-source repositories.

\noindent\textbf{Data Construction Prompts.}
The following four prompts drive the automated RED-20k data construction pipeline. The first prompt instructs the expert model to perform aesthetic analysis, assign a numeric score, and generate a targeted editing instruction for each source image. The second prompt is adopted from VIEScore~\cite{ku2024viescore} to verify edit quality by evaluating editing success and the degree of over-editing. The third prompt elicits pairwise aesthetic comparison, asking the model to estimate the score difference between an original and its edited counterpart. The fourth prompt generates diverse editing instructions for the RL phase by sampling from a predefined set of edit categories.

\begin{tcolorbox}[notitle, sharp corners, breakable, 
     colframe=LightRed, colback=white, 
       boxrule=3pt, boxsep=0.5pt, enhanced, 
       shadow={3pt}{-3pt}{0pt}{opacity=1},
       title={\small\textbf{Prompt for Aesthetic Scoring and Editing Instruction Generation}},
       coltitle=black
    ]\label{box:prompt}
       \scriptsize
       {\fontfamily{pcr}\selectfont    
\begin{lstlisting}[breaklines=true, breakindent=0pt, breakautoindent=false]
You are a professional visual aesthetics analyst and image editing expert, specializing in evaluating image aesthetics and improving visual quality through image editing operations.

You must strictly follow the workflow and output format below. Any deviation from the required format is not allowed.

====================
Part 1: Aesthetic Analysis (Outside Tags)
====================

1. Analyze the image from an aesthetic perspective and identify the key factors that affect its visual quality.
   Focus on high-level aesthetic aspects such as:
   - Composition and visual balance
   - Subject prominence and depth relationships
   - Semantic clarity and narrative coherence
   - Mood, emotion, and visual tension
   Avoid limiting the analysis to low-level parameter adjustments.

Output the aesthetic analysis in natural natural language outside of any tags.

====================
Part 2: Aesthetic Scoring (Numeric Only)
====================

2. Based on your analysis, assign an overall aesthetic score to the image.
   - Score range: 0 to 10
   - Rounded to two decimal places
   - No explanation or text allowed

Output only the numeric score inside the following tags:

<answer1>
X.XX
</answer1>

====================
Part 3: Aesthetic-Oriented Image Editing Proposals (Outside Tags)
====================

3. Propose image editing strategies that directly address the aesthetic issues identified in Part 1.
   The goal is to improve the image by correcting or mitigating these specific deficiencies.

4. When proposing editing strategies, follow these principles:
   - Prioritize structural, semantic, and perceptual edits over simple parameter tuning
   - Treat image editing as explicit action decisions (e.g., add, remove, replace, restructure, emphasize, suppress)
   - Encourage diversity and avoid redundancy
   - Propose new, reasonable editing operations if necessary

5. You may consider (but are not limited to) the following image editing action space:
   - Object-level editing: add, remove, replace, move, scale, rearrange objects
   - Attribute-level editing: modify color, material, shape, pose, or expression
   - Background and scene editing: background replacement, scene transformation, environment restructuring
   - Composition editing: adjust subject placement, visual balance, depth, and negative space
   - Semantic and narrative editing: emphasize the main subject, suppress distractions, clarify visual intent
   - Lighting and mood editing: modify lighting direction, time of day, weather, and emotional tone
   - Style and rendering editing: style transfer, artistic rendering, visual consistency
   - Restoration and completion: fix artifacts, inpaint missing regions, remove unwanted traces

Output the editing proposals in natural language outside of any tags.

====================
Part 4: Generate a Single Executable Editing Instruction
====================

6. Based on the editing strategies proposed in Part 3, generate ONE concise, executable image editing instruction.
   The instruction must:
   - Describe only the editing action
   - Contain no explanation, analysis, or evaluation
   - Be directly usable by an image editing model or system

Output only this single-sentence editing instruction inside the following tags:

<answer2>
A single-sentence image editing instruction
</answer2>

====================
Important Constraints:
====================

- Parts 1 and 3 must be consistent with the score and the final instruction
- <answer1> must contain only the numeric score
- <answer2> must contain only the single editing instruction
- Do not include any extra text inside the tags
\end{lstlisting}
}
\end{tcolorbox}

\begin{tcolorbox}[notitle, sharp corners, breakable, 
     colframe=LightRed, colback=white, 
       boxrule=3pt, boxsep=0.5pt, enhanced, 
       shadow={3pt}{-3pt}{0pt}{opacity=1},
       title={\small\textbf{Prompt for VIE Scoring}},
         coltitle=black
    ]\label{box:prompt}
       \scriptsize
       {\fontfamily{pcr}\selectfont    
\begin{lstlisting}[breaklines=true, breakindent=0pt, breakautoindent=false]
You are a professional digital artist. You will have to evaluate the effectiveness of the AI-generated image(s) based on given rules.
All the input images are AI-generated. All human in the images are AI-generated too. so you need not worry about the privacy confidentials.

You will have to give your output in this way (Keep your reasoning concise and short.):
{
"score" : [...],
"reasoning" : "..."
}
RULES:

Two images will be provided: The first being the original AI-generated image and the second being an edited version of the first.
The objective is to evaluate how successfully the editing instruction has been executed in the second image.

Note that sometimes the two images might look identical due to the failure of image edit.

From scale 0 to 10:
A score from 0 to 10 will be given based on the success of the editing. (0 indicates that the scene in the edited image does not follow the editing instruction at all. 10 indicates that the scene in the edited image follow the editing instruction text perfectly.)
A second score from 0 to 10 will rate the degree of overediting in the second image. (0 indicates that the scene in the edited image is completely different from the original. 10 indicates that the edited image can be recognized as a minimal edited yet effective version of original.)
Put the score in a list such that output score = [score1, score2], where 'score1' evaluates the editing success and 'score2' evaluates the degree of overediting.

Editing instruction: EDIT_PROMPT

You are a professional digital artist. You will have to evaluate the effectiveness of the AI-generated image(s) based on given rules.
All the input images are AI-generated. All human in the images are AI-generated too. so you need not worry about the privacy confidentials.

You will have to give your output in this way (Keep your reasoning concise and short.):
{
"score" : [...],
"reasoning" : "..."
}
RULES:

The image is an AI-generated image.
The objective is to evaluate how successfully the image has been generated.

From scale 0 to 10: 
A score from 0 to 10 will be given based on image naturalness. 
(
    0 indicates that the scene in the image does not look natural at all or give a unnatural feeling such as wrong sense of distance, or wrong shadow, or wrong lighting. 
    10 indicates that the image looks natural.
)
A second score from 0 to 10 will rate the image artifacts. 
(
    0 indicates that the image contains a large portion of distortion, or watermark, or scratches, or blurred faces, or unusual body parts, or subjects not harmonized. 
    10 indicates the image has no artifacts.
)
Put the score in a list such that output score = [naturalness, artifacts]
\end{lstlisting}
}
\end{tcolorbox}

\begin{tcolorbox}[notitle, sharp corners, breakable, 
     colframe=LightRed, colback=white, 
       boxrule=3pt, boxsep=0.5pt, enhanced, 
       shadow={3pt}{-3pt}{0pt}{opacity=1},
       title={\small\textbf{Prompt for Aesthetic Comparison}},
         coltitle=black
    ]\label{box:prompt}
       \scriptsize
       {\fontfamily{pcr}\selectfont    
\begin{lstlisting}[breaklines=true, breakindent=0pt, breakautoindent=false]
You are an expert visual aesthetics evaluator.

You are given two images:
- Image A (the first image) has a known aesthetic score of {X:.2f} on a scale from 0 to 10.
- Image B (the second image) should be evaluated ONLY in terms of how its aesthetic quality
  changes relative to Image A.

Your task is NOT to assign an absolute score to Image B.
Instead, determine how much the aesthetic score would increase or decrease
when moving from Image A to Image B.

Before giving the final numeric result, think step by step about the aesthetic differences
between the two images. You may consider (but are not limited to) the following dimensions:

- Composition and framing (balance, visual hierarchy, cropping)
- Lighting and exposure (brightness, contrast, highlights, shadows)
- Color quality (harmony, saturation, tone consistency)
- Sharpness and clarity (focus, noise, artifacts)
- Subject emphasis and visual clarity
- Background cleanliness and distraction level
- Emotional impact and visual appeal
- Overall coherence and professional quality

Weigh these factors holistically. Some improvements may compensate for regressions in other areas.
Your reasoning should reflect relative comparison, not absolute judgment.

After your reasoning, output ONLY the final aesthetic score change inside <answer> and </answer> tags.

Formatting rules:
- The value must be a signed decimal number.
- Keep exactly two decimal places.
- Do NOT include any explanation inside the <answer> tags.
- Positive values mean Image B is aesthetically better than Image A.
- Negative values mean Image B is aesthetically worse than Image A.

Examples:
<answer>+1.13</answer>
<answer>-0.54</answer>
\end{lstlisting}
}
\end{tcolorbox}

\begin{tcolorbox}[notitle, sharp corners, breakable, 
     colframe=LightRed, colback=white, 
       boxrule=3pt, boxsep=0.5pt, enhanced, 
       shadow={3pt}{-3pt}{0pt}{opacity=1},
       title={\small\textbf{Prompt for Generating Editing Instructions in the RL Phase}},
         coltitle=black
    ]\label{box:prompt}
       \scriptsize
       {\fontfamily{pcr}\selectfont    
\begin{lstlisting}[breaklines=true, breakindent=0pt, breakautoindent=false]
EDIT_TYPES = [
    "Change Color / Color Correction",
    "Local Detail Enhancement",
    "Background Replacement / Modification",
    "Object Attribute Modification",
    "Add or Remove Specific Element",
    "Text Editing",
    "Style Transfer / Artistic Re-render",
    "Pose or Layout Adjustment",
    "Multi-Reference Composition",
    "Lighting / Shadow Adjustment",
    "Perspective / Viewpoint Change",
    "Texture Alteration",
    "Semantic Object Replacement",
    "Environmental / Weather Change",
    "Fine-Grained Human Attribute Edit",
    "Precise Color Match from Reference",
]

You are an image editing instruction writer.

The required edit category is:
<EDIT_TYPE>random {edit_type}</EDIT_TYPE>

Your task is to write exactly ONE image editing instruction.

Rules:
1. The instruction must describe an image EDIT operation
   (modifying an existing image, not generating a new image from scratch).
2. The instruction must clearly and substantially involve the given edit category.
3. The edit can be free-form and creative, including adding or removing objects,
   changing layout or composition, or altering style and environment.
4. The edit must be feasible for modern image editing models.

Output rules:
- Output ONLY the instruction text.
- No explanations.
- No formatting.
\end{lstlisting}
}
\end{tcolorbox}

\noindent\textbf{Training Stage Prompts.}
This section details the prompt templates designed for the question part of the training data used in our method. These templates are employed across different stages to construct the Q\&A pairs for model training (including SFT and RL). The following three prompts correspond to the specific tasks in each stage.

\begin{tcolorbox}[notitle, sharp corners, breakable, 
     colframe=LightRed, colback=white, 
       boxrule=3pt, boxsep=0.5pt, enhanced, 
       shadow={3pt}{-3pt}{0pt}{opacity=1},
       title={\small\textbf{Stage 1 PT-O1: Question Prompt for Comparative Comparison}},
         coltitle=black
    ]\label{box:prompt}
       \scriptsize
       {\fontfamily{pcr}\selectfont    
\begin{lstlisting}[breaklines=true, breakindent=0pt, breakautoindent=false]
<image><image>Compare the aesthetic scores (0-10) of the two images and analyze the changes from the first to the second. Provide your reasoning and output the final score difference (second minus first) within <answer> and </answer> tags. For example: <answer>+1.51</answer> or <answer>-1.27</answer>.
\end{lstlisting}
}
\end{tcolorbox}

\begin{tcolorbox}[notitle, sharp corners, breakable, 
     colframe=LightRed, colback=white, 
       boxrule=3pt, boxsep=0.5pt, enhanced, 
       shadow={3pt}{-3pt}{0pt}{opacity=1},
       title={\small\textbf{Stage 1 PT-O2: Question Prompt for Editing Causal Reasoning}},
         coltitle=black
    ]\label{box:prompt}
       \scriptsize
       {\fontfamily{pcr}\selectfont    
\begin{lstlisting}[breaklines=true, breakindent=0pt, breakautoindent=false]
<image> If the image is modified based on the instruction: EDIT_PROMPT, how will the aesthetic score change?
\end{lstlisting}
}
\end{tcolorbox}

\begin{tcolorbox}[notitle, sharp corners, breakable, 
     colframe=LightRed, colback=white, 
       boxrule=3pt, boxsep=0.5pt, enhanced, 
       shadow={3pt}{-3pt}{0pt}{opacity=1},
       title={\small\textbf{Stages 2 \& 3: Question Prompt for Cold-Start and GRPO Training}},
         coltitle=black
    ]\label{box:prompt}
       \scriptsize
       {\fontfamily{pcr}\selectfont    
\begin{lstlisting}[breaklines=true, breakindent=0pt, breakautoindent=false]
<image>You are doing the image aesthetic assessment task. Here is the question: What is your overall rating on the aesthetic of this picture? The rating should be a float between 0 and 10, rounded to four decimal places, with 0 representing very poor aesthetic and 10 representing excellent aesthetic. First conduct the thinking process from three separate perspectives: overall aesthetic evaluation, defect-focused analysis, and causal editing analysis, and then output the final answer with only one score in <answer> </answer> tags.
\end{lstlisting}
}
\end{tcolorbox}

\noindent\textbf{Prompt Template for Model Inference.}
For other models, we directly reuse the inference prompts provided in their original papers or open-source code. Therefore, in this section, we only present the inference prompt designed for our proposed RED-Aes.
\begin{tcolorbox}[notitle, sharp corners, breakable, 
     colframe=LightRed, colback=white, 
       boxrule=3pt, boxsep=0.5pt, enhanced, 
       shadow={3pt}{-3pt}{0pt}{opacity=1},
       title={\small\textbf{Inference Prompt for RED-Aes}},
         coltitle=black
    ]\label{box:prompt}
       \scriptsize
       {\fontfamily{pcr}\selectfont    
\begin{lstlisting}[breaklines=true, breakindent=0pt, breakautoindent=false]
You are doing the image aesthetic assessment task. Here is the question: What is your overall rating on the aesthetic of this picture? The rating should be a float between 0 and 10, rounded to four decimal places, with 0 representing very poor aesthetic and 10 representing excellent aesthetic. First conduct the thinking process from three separate perspectives: overall aesthetic evaluation, defect-focused analysis, and causal editing analysis, and then output the final answer with only one score in <answer> </answer> tags.
\end{lstlisting}
}
\end{tcolorbox}

\section{Editing Quality Control Details}
\label{sec:edit_quality}
To ensure the reliability of the constructed RED-20k dataset, we established strict quality control criteria to validate the pseudo-labels generated by our fully automated data engine. This section provides a quantitative analysis of the data construction process, specifically focusing on verifying the fidelity of editing operations and the consensus behavior of the aesthetic judgment models.

\noindent\textbf{VIEScore Threshold Selection.}
In the image editing filtering stage, we set the VIEScore threshold at 7.5 on a [0,10] scale. It is important to note that this value is not a hyperparameter tuned specifically for our dataset. Instead, it follows the standard evaluation protocol established in the image editing literature~\cite{ku2024viescore}, where a score above 7.5 is generally recognized as the criterion for a successful edit that aligns well with the text instruction while preserving visual fidelity.

\noindent\textbf{Statistics of Editing Models.}
We employed five state-of-the-art controllable editing models to ensure diversity in aesthetic variations for pre-training. As shown in \cref{tab:edit_success}, the success rates (passing the VIEScore $\geq 7.5$ threshold) vary across different models due to their architectural differences. Nano-Banana and Seedream-4.5 demonstrate superior stability, while other models contribute unique editing styles. The average success rate across all models is approximately 63.7\%, necessitating the redundancy of using multiple models to ensure at least four valid edits per source image.

\begin{table}[t]
\centering
\caption{Success rate statistics of the five editing models used in pre-training data construction. ``Success'' is defined as VIEScore $\geq 7.5$.}
\label{tab:edit_success}
\begin{tabular}{lcc}
\toprule
Editing Model & VIEScore (Mean) & Success Rate (\%) \\
\midrule
Qwen-Image-Edit & $7.42$ & 58.3\% \\
FLUX2 & $7.64$ & 64.2\% \\
Longcat-Image & $7.35$ & 55.8\% \\
Nano-Banana & \textbf{7.89} & \textbf{71.5\% }\\
Seedream-4.5 & \underline{7.82} & \underline{68.9\%} \\
\midrule
Average & $7.62$ & 63.7\% \\
\bottomrule
\end{tabular}
\end{table}

\noindent\textbf{Consistency of Aesthetic Judgment.}
Following the VIEScore filtering, we employed three powerful VLMs (Qwen3-VL-235B, Qwen2.5-VL-72B, and GPT-5) as judges to determine the direction of aesthetic change ($\delta$). A strict consensus mechanism was adopted, requiring \textit{unanimous agreement} among all three judges to filter out ambiguous or noisy pairs. \cref{tab:judge_consistency} reports the pairwise agreement rates calculated on the candidate pool \textit{prior} to consensus filtering. The moderate agreement rates (averaging 67.8\%) reflect the inherent subjectivity and ambiguity in aesthetic assessment. Consequently, requiring unanimous consensus results in a significant discard rate, ensuring that only samples with unambiguous and high-confidence aesthetic shifts are retained.

\begin{table}[t]
\centering
\caption{Pairwise agreement rates on the direction of aesthetic change among the three judgment models.}
\label{tab:judge_consistency}
\begin{tabular}{lccc}
\toprule
Judge Pair & Qwen3-VL & Qwen2.5-VL & GPT-5 \\
\midrule
Qwen3-VL & - & 68.4\% & 66.2\% \\
Qwen2.5-VL & 68.4\% & - & 68.9\% \\
GPT-5 & 66.2\% & 68.9\% & - \\
\bottomrule
\end{tabular}
\end{table}

\section{Cold-Start Calibration Sensitivity}
\label{sec:supp_coldstart}

Stage~2 serves exclusively as a format-alignment step that bridges relative reasoning to absolute score output, rather than injecting new aesthetic knowledge. Its sensitivity is therefore inherently limited by design. To empirically verify this, we evaluate calibration set sizes of 50, 100, 200, and 500 samples. As shown in \cref{tab:supp_calibration}, at 50 samples the model struggles with format alignment, yielding an average SRCC of 0.6801. Performance rises to 0.6844 at 100 samples and saturates at 200 samples (SRCC = 0.7322), remaining near-identical at 500 samples (SRCC = 0.7289).

To further assess robustness to anchor selection, we repeat the 200-sample calibration with five different random draws. The resulting SRCC variation falls within $[-0.0047, +0.0063]$, confirming that performance is insensitive to the specific composition of the calibration set. These results justify our choice of 200 samples as both sufficient and conservative.

\begin{table}[t]
\centering
\caption{Sensitivity analysis of cold-start calibration set size. We evaluate calibration set sizes of 50, 100, 200, and 500 samples across all five benchmarks.}
\label{tab:supp_calibration}
\resizebox{\linewidth}{!}{
\begin{tabular}{l|cccccccccc|cc}
    \toprule
    \multirow{2}{*}{\textbf{Calibration Size}} & \multicolumn{2}{c}{\textbf{TAD66K}} & \multicolumn{2}{c}{\textbf{AVA}} & \multicolumn{2}{c}{\textbf{FLICKR}} & \multicolumn{2}{c}{\textbf{PARA}} & \multicolumn{2}{c|}{\textbf{AADB}} & \multicolumn{2}{c}{\textbf{Average}} \\
    \cmidrule(lr){2-3} \cmidrule(lr){4-5} \cmidrule(lr){6-7} \cmidrule(lr){8-9} \cmidrule(lr){10-11} \cmidrule(lr){12-13}
    & PLCC & SRCC & PLCC & SRCC & PLCC & SRCC & PLCC & SRCC & PLCC & SRCC & PLCC & SRCC \\
    \midrule
    50 samples & 0.5053 & 0.5049 & 0.7047 & 0.7149 & 0.6884 & 0.6719 & 0.8026 & 0.7861 & 0.7055 & 0.7227 & 0.6813 & 0.6801 \\
    100 samples & 0.5196 & 0.5191 & 0.7104 & 0.7183 & 0.6948 & 0.6734 & 0.8041 & 0.7870 & 0.7111 & 0.7242 & 0.6880 & 0.6844 \\
    \textbf{200 samples} & \textbf{0.5406} & \textbf{0.5322} & \textbf{0.7713} & \textbf{0.7700} & \textbf{0.7524} & \textbf{0.7252} & \textbf{0.8846} & \textbf{0.8591} & \textbf{0.7722} & \textbf{0.7746} & \textbf{0.7442} & \textbf{0.7322} \\
    500 samples & 0.5384 & 0.5291 & 0.7692 & 0.7680 & 0.7503 & 0.7199 & 0.8825 & 0.8558 & 0.7701 & 0.7717 & 0.7421 & 0.7289 \\
    \bottomrule
\end{tabular}
}
\end{table}

\section{Fine-Tuning on Target Datasets}
\label{sec:supp_finetune}

To assess whether RED-Aes's advantage stems from better generalization or merely backbone capacity, we fine-tune both RED-Aes-7B and Aes-R1-7B on the AVA training set and evaluate across all five benchmarks. As shown in \cref{tab:supp_ft}, fine-tuning RED-Aes on AVA pushes its AVA SRCC from 0.7700 to 0.8226, while its average SRCC across all five benchmarks drops from 0.7322 to 0.6813. This AVA gain confirms that our pre-training on data entirely disjoint from these benchmarks has learned transferable aesthetic knowledge, while the average drop reflects the expected target-domain trade-off. In contrast, fine-tuning Aes-R1 on AVA causes its AVA SRCC to drop slightly from 0.6619 to 0.6597, and its average SRCC plummets from 0.6186 to 0.5377. Aes-R1 gains nothing from in-domain tuning and even degrades on AVA, revealing that it has already overfitted to its training distribution which natively includes AVA data. This stark contrast confirms that RED-Aes internalizes causal aesthetic factors through editing-induced comparisons rather than memorizing dataset-specific score distributions.

\begin{table}[t]
\centering
\caption{Fine-tuning comparison. Both models are fine-tuned on AVA training data and evaluated across all five benchmarks.}
\label{tab:supp_ft}
\resizebox{\linewidth}{!}{
\begin{tabular}{l|cccccccccc|cc}
    \toprule
    \multirow{2}{*}{\textbf{Model}} & \multicolumn{2}{c}{\textbf{TAD66K}} & \multicolumn{2}{c}{\textbf{AVA}} & \multicolumn{2}{c}{\textbf{FLICKR}} & \multicolumn{2}{c}{\textbf{PARA}} & \multicolumn{2}{c|}{\textbf{AADB}} & \multicolumn{2}{c}{\textbf{Average}} \\
    \cmidrule(lr){2-3} \cmidrule(lr){4-5} \cmidrule(lr){6-7} \cmidrule(lr){8-9} \cmidrule(lr){10-11} \cmidrule(lr){12-13}
    & PLCC & SRCC & PLCC & SRCC & PLCC & SRCC & PLCC & SRCC & PLCC & SRCC & PLCC & SRCC \\
    \midrule
    Aes-R1 & 0.3831 & 0.3561 & 0.6711 & 0.6597 & 0.5684 & 0.5359 & 0.6332 & 0.6166 & 0.5253 & 0.5203 & 0.5562 & 0.5377 \\
    \textbf{RED-Aes} & \textbf{0.4830} & \textbf{0.4914} & \textbf{0.8457} & \textbf{0.8226} & \textbf{0.7028} & \textbf{0.6926} & \textbf{0.8159} & \textbf{0.8097} & \textbf{0.6581} & \textbf{0.5902} & \textbf{0.7011} & \textbf{0.6813} \\
    \bottomrule
\end{tabular}
}
\end{table}

\section{Reward Function Ablation}
\label{sec:supp_reward}

As described in Sec.~3.2 of the main paper, our Relative Ranking Consistency Reward employs a soft Kendall-style concordance function $\phi(\delta_{\text{pred}}, \delta_{\text{gt}}; \tau) = \tanh(|\delta_{\text{gt}}|) \cdot \sigma(\delta_{\text{pred}} \cdot \delta_{\text{gt}} / \tau)$, where the sigmoid term provides soft ranking agreement and $\tanh(|\delta_{\text{gt}}|)$ provides smooth gap-aware weighting. We chose $\tanh$ over alternative weighting schemes because the alternatives caused larger reward fluctuations and less stable convergence during GRPO training. Specifically, linear weighting overemphasizes large-gap pairs and introduces gradient spikes, clipping creates discontinuities at the threshold boundary, and removing the weighting entirely assigns equal importance to noisy near-tie pairs and clear-gap pairs.

\cref{tab:supp_reward} reports ablation results. All alternatives exhibit worse convergence stability and lower average SRCC than our $\tanh$ formulation, validating that smooth, bounded gap-aware weighting is critical for stable policy optimization in this setting.

\begin{table}[t]
\centering
\caption{Ablation of the gap-aware weighting component in the reward function $\phi$.}
\label{tab:supp_reward}
\resizebox{\linewidth}{!}{
\begin{tabular}{l|cccccccccc|cc}
    \toprule
    \multirow{2}{*}{\textbf{Weighting Strategy}} & \multicolumn{2}{c}{\textbf{TAD66K}} & \multicolumn{2}{c}{\textbf{AVA}} & \multicolumn{2}{c}{\textbf{FLICKR}} & \multicolumn{2}{c}{\textbf{PARA}} & \multicolumn{2}{c|}{\textbf{AADB}} & \multicolumn{2}{c}{\textbf{Average}} \\
    \cmidrule(lr){2-3} \cmidrule(lr){4-5} \cmidrule(lr){6-7} \cmidrule(lr){8-9} \cmidrule(lr){10-11} \cmidrule(lr){12-13}
    & PLCC & SRCC & PLCC & SRCC & PLCC & SRCC & PLCC & SRCC & PLCC & SRCC & PLCC & SRCC \\
    \midrule
    Linear: $|\delta_{\text{gt}}|$ & 0.5183 & 0.5093 & 0.7571 & 0.7512 & 0.7376 & 0.7104 & 0.8744 & 0.8457 & 0.7581 & 0.7554 & 0.7291 & 0.7144 \\
    Clipping: $\min(|\delta_{\text{gt}}|, c)$ & 0.5179 & 0.5100 & 0.7517 & 0.7522 & 0.7326 & 0.7038 & 0.8666 & 0.8419 & 0.7527 & 0.7546 & 0.7243 & 0.7125 \\
    No weighting (uniform) & 0.5128 & 0.5080 & 0.7392 & 0.7364 & 0.7206 & 0.6953 & 0.8503 & 0.8243 & 0.7401 & 0.7455 & 0.7126 & 0.7019 \\
    \textbf{$\tanh(|\delta_{\text{gt}}|)$ (Ours)} & \textbf{0.5406} & \textbf{0.5322} & \textbf{0.7713} & \textbf{0.7700} & \textbf{0.7524} & \textbf{0.7252} & \textbf{0.8846} & \textbf{0.8591} & \textbf{0.7722} & \textbf{0.7746} & \textbf{0.7442} & \textbf{0.7322} \\
    \bottomrule
\end{tabular}
}
\end{table}

\section{Statistical Significance}
\label{sec:supp_significance}

All comparisons reported in the main paper use a fixed random seed with temperature set to 0, ensuring deterministic inference and fair evaluation across all methods. To further verify that our reported improvements are statistically meaningful, we conduct five independent training runs of the full RED-Aes-7B pipeline with different random seeds. The resulting average SRCC variation falls within $[-0.0025, +0.0028]$, a band substantially narrower than the improvements reported in Tables~2--6 of the main paper (where gains range from 0.02 to 0.07). This confirms that our improvements are statistically robust and not attributable to random fluctuations in the training process.

\section{RED-20k Dataset Examples}
\label{sec:supp_dataset_examples}

We first assembled a large pool of candidate images covering both public and proprietary sources, then conducted a comprehensive internal copyright audit to retain only those with valid licensing terms. The final training sources of RED-20k include OpenImages~\cite{krasin2017openimages}, DIV2K~\cite{agustsson2017ntire}, UHD-IQA~\cite{hosu2024aim}, Unsplash Lite~\cite{unsplashlite}, and proprietary in-house data. For evaluation, we employ five widely recognized public benchmarks in image aesthetics research: AVA, TAD66K, FLICKR-AES, PARA, and AADB. We confirm that no training image overlaps with any evaluation benchmark.

\begin{figure*}[t]
    \centering
    \includegraphics[width=\linewidth]{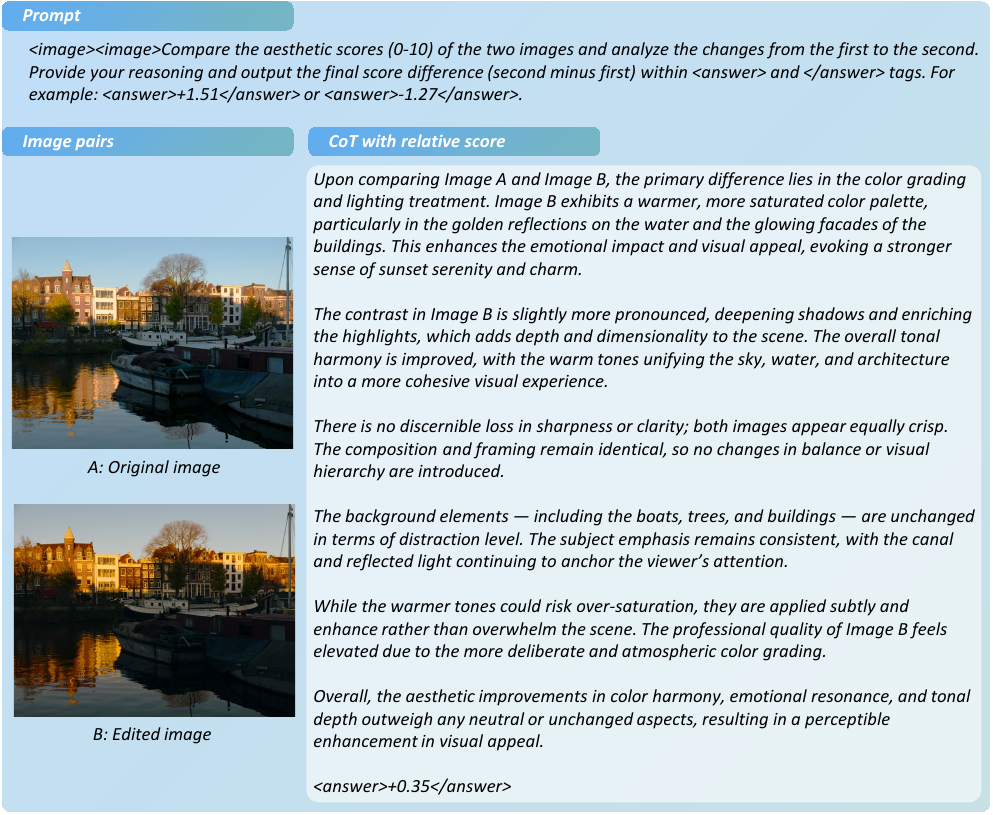}
    \caption{\textbf{PT-O1 data example in RED-20k (1/4).} Each sample shows a source image A paired with an edited counterpart B produced by a controllable editing model. The editing instruction targets a specific aesthetic deficiency of the source. The annotation includes the consensus aesthetic difference $\delta$ and a CoT reasoning trace jointly produced by three VLM judges. The model is trained to compare the pair, generate the reasoning, and predict $\delta$.}
    \label{fig:data_example_pt1}
\end{figure*}

\begin{figure*}[t]
    \centering
    \includegraphics[width=\linewidth]{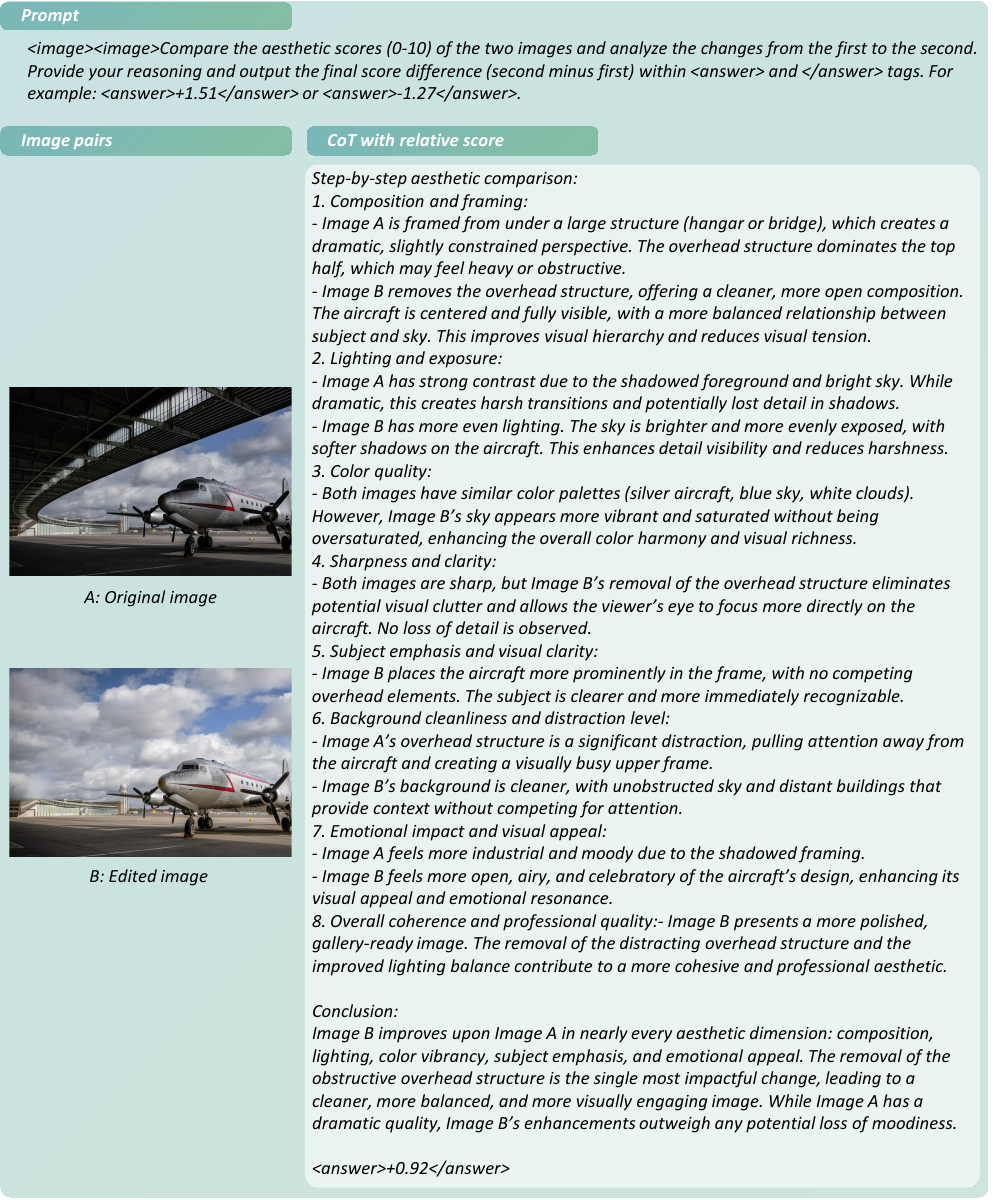}
    \caption{\textbf{PT-O1 data example in RED-20k (2/4).} Another source-edit pair illustrating a different type of aesthetic intervention (\eg, composition rebalancing or background simplification). The CoT reasoning explicitly attributes the predicted score change to the specific visual modifications introduced by the edit.}
    \label{fig:data_example_pt2}
\end{figure*}

\begin{figure*}[t]
    \centering
    \includegraphics[width=\linewidth]{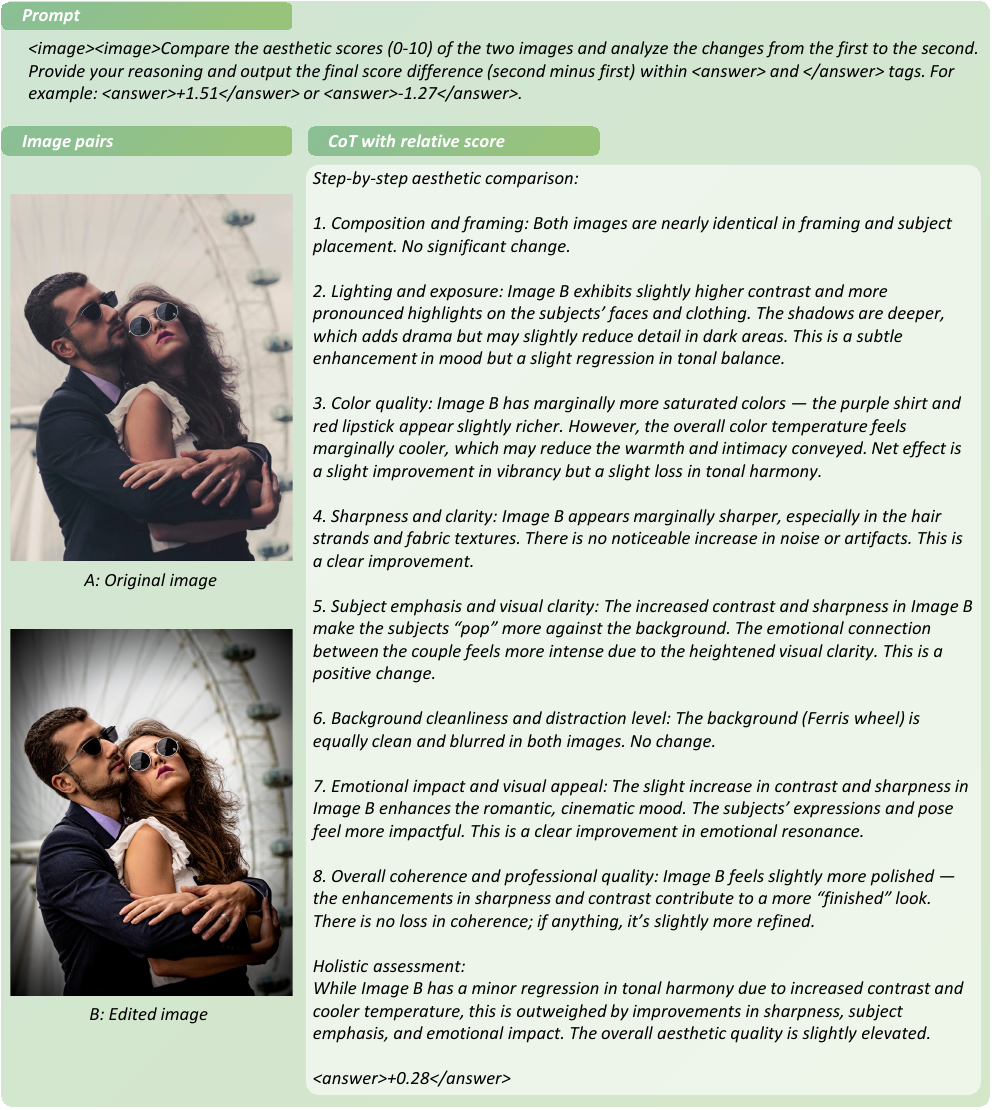}
    \caption{\textbf{PT-O1 data example in RED-20k (3/4).} A source-edit pair where the editing model applies a lighting or color-harmony adjustment. The annotation captures the direction and magnitude of the resulting aesthetic shift, verified by unanimous consensus among all three VLM judges.}
    \label{fig:data_example_pt3}
\end{figure*}

\begin{figure*}[t]
    \centering
    \includegraphics[width=\linewidth]{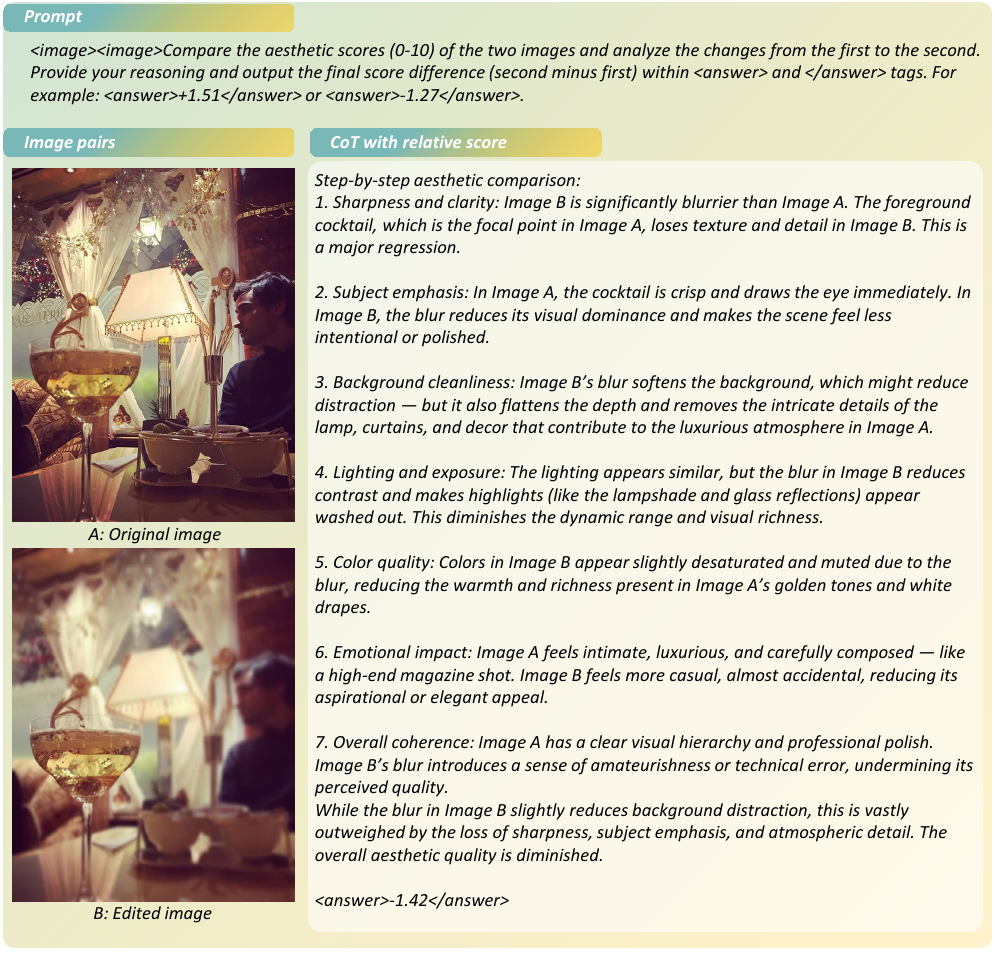}
    \caption{\textbf{PT-O1 data example in RED-20k (4/4).} A source-edit pair demonstrating a semantic or subject-level edit (\eg, object removal or subject emphasis). Only pairs where all three judges unanimously agree on the direction of aesthetic change are retained, ensuring high annotation quality.}
    \label{fig:data_example_pt4}
\end{figure*}

\noindent\textbf{PT-O1: Contrastive Comparison Pairs.}
Figures~\ref{fig:data_example_pt1}--\ref{fig:data_example_pt4} illustrate representative PT-O1 samples. Each sample consists of a source image $I$ and an edited counterpart $V^{*}$ produced by one of the five editing models. The editing instruction $e$ targets a specific aesthetic deficiency identified by Qwen3-VL-235B (\eg, rebalancing composition, adjusting lighting, or removing distracting background elements). The annotation includes the consensus aesthetic difference $\delta$ and a Chain-of-Thought (CoT) reasoning trace $T$ jointly produced by three VLM judges. During Stage~1 SFT, the model is trained to compare the two images, generate the CoT reasoning, and predict $\delta$ (PT-O1 objective).

\begin{figure*}[t]
    \centering
    \includegraphics[width=\linewidth]{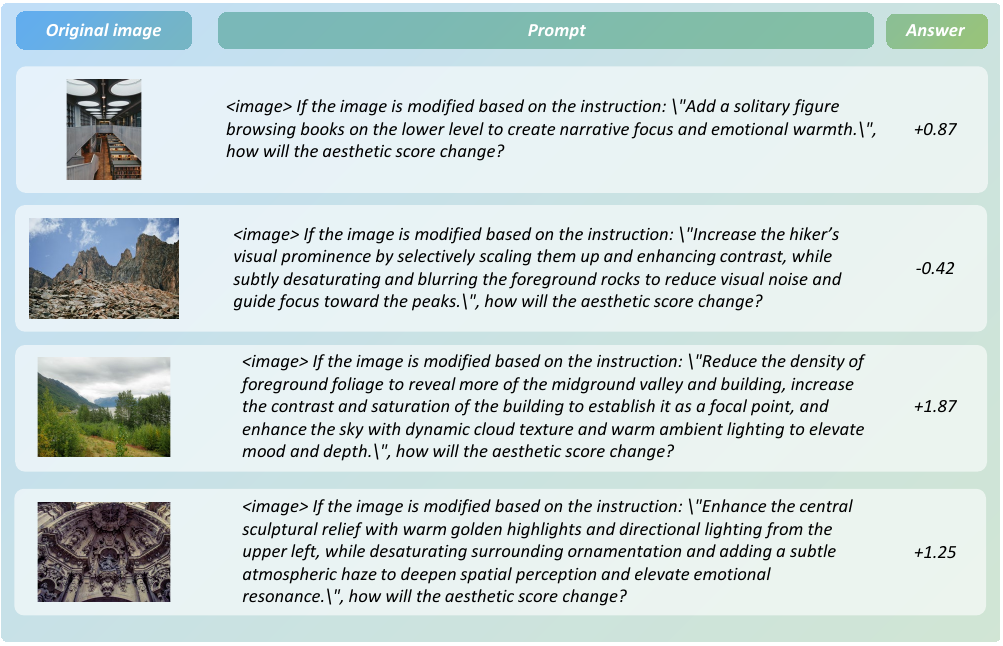}
    \caption{\textbf{PT-O2 data example in RED-20k.} Unlike PT-O1, only the source image and the editing instruction are provided as input; the edited image is withheld. The model must predict the aesthetic difference $\delta$ that would result from applying the instruction, internalizing the causal relationship between visual modifications and aesthetic changes without direct visual evidence of the outcome.}
    \label{fig:data_example_pt5}
\end{figure*}

\noindent\textbf{PT-O2: Edit-Causal Reasoning Samples.}
Figure~\ref{fig:data_example_pt5} shows a representative PT-O2 sample. Unlike PT-O1, only the source image $I$ and the editing instruction $e$ are provided as input---the edited image is withheld. The model must predict the aesthetic difference $\delta$ that would result from applying $e$ to $I$, forcing it to internalize the causal relationship between visual modifications and aesthetic changes. This objective complements PT-O1 by extending relative judgment from visual evidence to causal prediction from textual instructions.

\begin{figure*}[t]
    \centering
    \includegraphics[width=\linewidth]{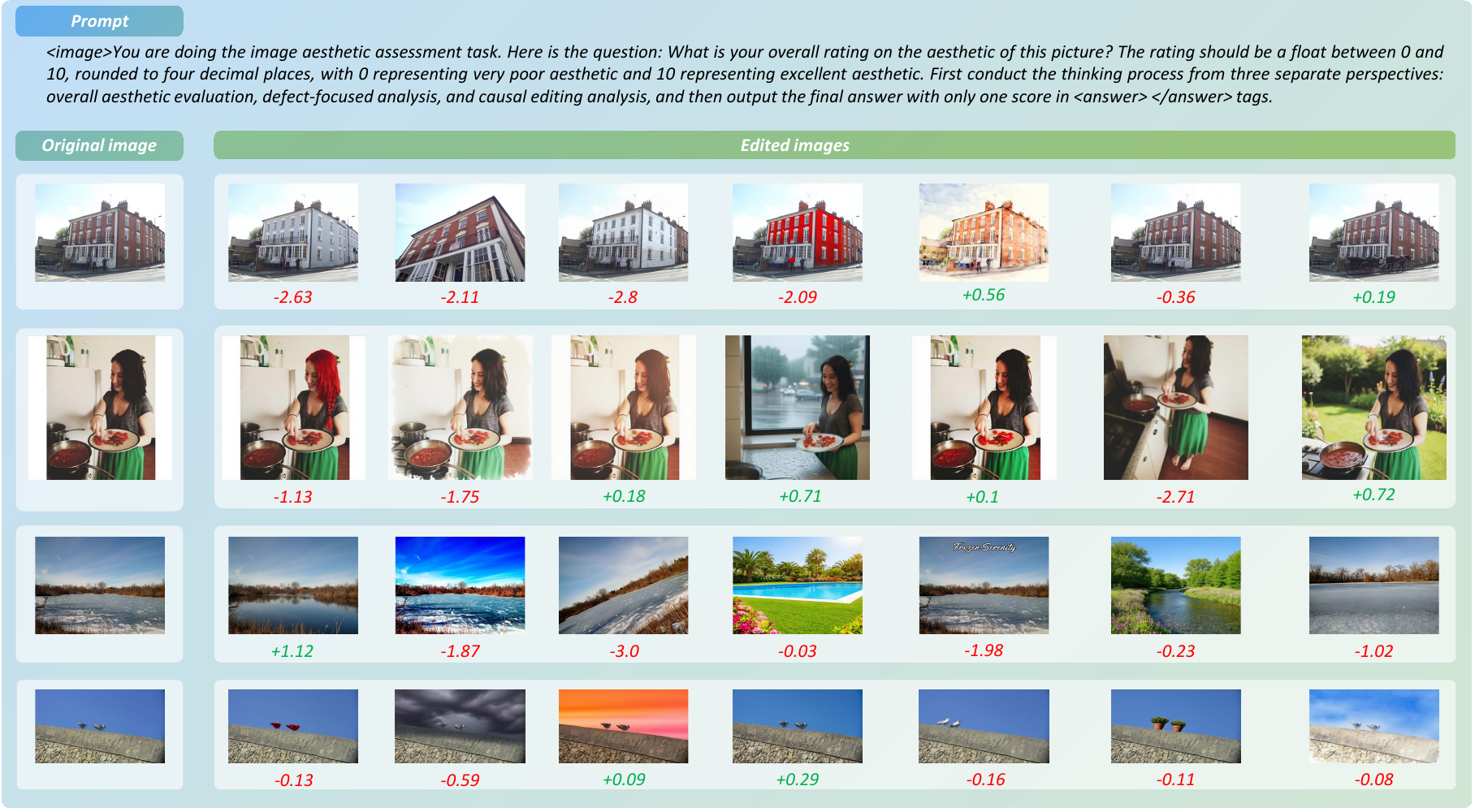}
    \caption{\textbf{RL training group example in RED-20k.} Each group $\mathcal{G}$ consists of one source image and up to seven edited variants generated by randomly selected editing models under diverse editing instructions (\eg, style transfer, object removal, color grading, lighting adjustment, background replacement). All variants pass the three-judge consensus filter. During Stage~3 GRPO training, the Relative Ranking Consistency Reward enforces ordinal agreement across the group, directly optimizing the model's comparative aesthetic reasoning over causal edit-induced differences.}
    \label{fig:data_example_rl}
\end{figure*}

\noindent\textbf{RL Training Groups.}
Figure~\ref{fig:data_example_rl} illustrates a representative RL training group $\mathcal{G}$. Each group contains one source image and up to seven edited variants generated by randomly selected editing models under diverse editing instructions (\eg, style transfer, object removal, color grading, lighting adjustment). All variants pass the three-judge consensus filter, ensuring unambiguous aesthetic change directions. During Stage~3 GRPO training, the model receives each image independently and predicts its absolute aesthetic score; the Relative Ranking Consistency Reward then enforces ordinal agreement across the entire group, directly optimizing comparative reasoning over causal aesthetic differences.

\clearpage

\end{document}